\documentclass[acmsmall]{acmart}

\usepackage{tabularx}
\usepackage{soul}
\usepackage{stfloats}
\usepackage{comment}
\usepackage{adjustbox}
\usepackage{titlesec}

\usepackage{color}
\usepackage{graphicx}
\usepackage{ascmac}
\usepackage{CJKutf8}
\usepackage{bm}
\usepackage{bibentry}
\usepackage{booktabs}
\usepackage{subfig}
\usepackage{tcolorbox}
\usepackage{colortbl}

\usepackage{booktabs}
\usepackage{url}
\usepackage{multirow}
\usepackage{xcolor} 
\usepackage{diagbox}
\usepackage{multicol}
\usepackage[ruled,linesnumbered]{algorithm2e}
\usepackage{inputenc}
\usepackage{comment}
\usepackage{colortbl}

\AtBeginDocument{%
  \providecommand\BibTeX{{%
    \normalfont B\kern-0.5em{\scshape i\kern-0.25em b}\kern-0.8em\TeX}}}

\setcopyright{acmcopyright}
\acmJournal{TALLIP}
\begin{document}
\title{Linguistically-driven Multi-task Pre-training for Low-resource Neural Machine Translation}


\author{Zhuoyuan Mao}
\email{zhuoyuanmao@nlp.ist.i.kyoto-u.ac.jp}
\affiliation{\institution{Graduate School of Informatics, Kyoto University} \country{Japan}}

\author{Chenhui Chu}
\email{chu@nlp.ist.i.kyoto-u.ac.jp}
\affiliation{\institution{Graduate School of Informatics, Kyoto University} \country{Japan}}

\author{Sadao Kurohashi}
\email{kuro@nlp.ist.i.kyoto-u.ac.jp}
\affiliation{\institution{Graduate School of Informatics, Kyoto University} \country{Japan}}

\renewcommand{\shortauthors}{Zhuoyuan Mao, Chenhui Chu, and Sadao Kurohashi}


\begin{abstract}
%In the present study, we employ one of the state-of-the-art sequence-to-sequence pre-training methods for low-resource neural machine translation (NMT), masked sequence-to-sequence (MASS), as the main baseline, and propose novel pre-training alternatives to MASS: 
In the present study, we propose novel sequence-to-sequence pre-training objectives for low-resource machine translation (NMT): Japanese-specific sequence to sequence (JASS) for language pairs involving Japanese as the source or target language, and English-specific sequence to sequence (ENSS) for language pairs involving English. JASS focuses on masking and reordering Japanese linguistic units known as bunsetsu, whereas ENSS is proposed based on phrase structure masking and reordering tasks. Experiments on ASPEC Japanese--English \& Japanese--Chinese, Wikipedia Japanese--Chinese, News English--Korean corpora demonstrate that JASS and ENSS outperform MASS and other existing language-agnostic pre-training methods by up to +2.9 BLEU points for the Japanese--English tasks, up to +7.0 BLEU points for the Japanese--Chinese tasks and up to +1.3 BLEU points for English--Korean tasks. Empirical analysis, which focuses on the relationship between individual parts in JASS and ENSS, reveals the complementary nature of the subtasks of JASS and ENSS. Adequacy evaluation using LASER, human evaluation, and case studies reveals that our proposed methods significantly outperform pre-training methods without injected linguistic knowledge and they have a larger positive impact on the adequacy as compared to the fluency. We release codes here: \url{https://github.com/Mao-KU/JASS/tree/master/linguistically-driven-pretraining}.
\end{abstract}




\maketitle

\section{Introduction}
Neural machine translation (NMT)~\cite{DBLP:journals/corr/BahdanauCB14:original,DBLP:journals/corr/SutskeverVL14} can achieve state-of-the-art performance when large parallel corpora are available for training. However, this prerequisite for parallel corpora limits its usefulness for several language pairs, such as Japanese, Chinese, and Korean, along with domains (history and COVID) for which such large corpora do not exist. Often, these resource-poor language pairs consist of languages that have resource-rich monolingual corpora. Therefore, it is possible to compensate for the lack of parallel corpora by leveraging large monolingual corpora. One popular approach for this is data augmentation, for instance, through back-translation~\cite{sennrich-etal-2016-improving,hoang-etal-2018-iterative}. Another approach involves pre-training the NMT model on tasks that only require monolingual corpora~\cite{qi-etal-2018-pre,song2019mass}.

As a promising technique for leveraging monolingual corpora, pre-training has experienced a surge in popularity in NLP ever since models such as BERT~\cite{devlin-etal-2019-bert} achieved state-of-the-art results in text understanding. However, BERT-like models were not designed to be used for NMT in the sense that they are essentially techniques for pre-training encoders, but not sequence-to-sequence models. To address this,~\citet{song2019mass},~\citet{lewis-etal-2020-bart} and~\citet{DBLP:journals/tacl/LiuGGLEGLZ20} recently proposed self-supervised language-agnostic pre-training methods, which are sequence-to-sequence pre-training tasks for NMT, have achieved new state-of-the-art results in low-resource scenarios.

Languages that are sufficiently ``rich" to have large monolingual corpora often have available tools for linguistic analysis. Meanwhile, usually a low-resource language pair is composed by a resource-rich language and a low-resource language and the linguistic knowledge of the resource-rich language can be easily extracted. In addition, studies such as~\citet{sennrich-haddow-2016-linguistic} and~\citet{rudramurthy19} have demonstrated that linguistic knowledge can improve NMT without using additional corpora. Therefore, it is natural to use monolingual corpora and linguistic tools in bilingual low-resource scenarios. However, the manner in which linguistic knowledge should be provided is not always clear, because NMT models are implemented in an end-to-end scheme. From a technical perspective, it is practical to extract linguistic features on the monolingual side. Therefore, monolingual pre-training provides an ideal framework for leveraging monolingual corpora and injecting linguistic information.

In~\citet{mao-etal-2020-jass}, we proposed a linguistically motivated pre-training approach known as Japanese-specific sequence-to-sequence (JASS), which was inspired by masked sequence-to-sequence pre-training (MASS), but focused on syntactic analysis obtained by using a parser. Particularly, we added syntactic constraints to the sentence-masking process of the MASS to obtain the bunsetsu-based MASS (BMASS) task.\footnote{For BMASS, bunsetsus are used as syntactic spans, which is the elementary syntactic component of Japanese. It can be extracted using the KNP.~\cite{kurohashi--EtAl:1994,jumanpp2}} We also proposed the Bunsetsu reordering-based sequence-to-sequence (BRSS), which is a linguistically motivated reordering task. Several previous studies~\cite{lewis-etal-2020-bart,DBLP:journals/jmlr/RaffelSRLNMZLL20} have provided evidence that ``multi-task" pre-training that combines various styles of self-supervised training tasks results in significantly superior results for NMT. We proposed JASS based on a combination of the above-mentioned two tasks and it is tailored for NMT involving Japanese. 

In contrast, in this study, we also propose linguistically-driven pre-training methods for English to leverage linguistic-specific information in the pre-training phase.\footnote{Although some language pairs involving English are middle- or high-resource scenarios (parallel corpora size over 100k), we deem that it is worth proposing methods for English because a large number of low-resource language pairs involving English are still present.} They are referred to as phrase structure-based MASS (PMASS) \& head finalization-based sequence-to-sequence (HFSS), and their combination is denoted as English-specific sequence-to-sequence (ENSS).\footnote{Head finalization~\cite{isozaki-etal-2010-head} is the technique used to reorder sentences in SVO language to be SOV-like sentences.}  Moreover, unlike the proposed methods for Japanese, the proposed methods for English can be transplanted onto any SVO language. Thus, our proposed ENSS and JASS can be applied to any translation pair involving English or Japanese.\footnote{According to the reordering task we proposed (specifically, BRSS and HFSS), more significant improvements are expected to observe on English--SOV language or Japanese--SVO language.}

We experimented with ASPEC Japanese--English \& Japanese--Chinese~ \cite{nakazawa-etal-2016-aspec}, Wikipedia Japanese--Chinese~ \cite{chu-etal-2014-constructing,DBLP:journals/talip/ChuNK16}, and News English--Korean~\cite{park-etal-2016-korean} in various pre-training settings for JASS and ENSS.\footnote{In~\citet{mao-etal-2020-jass}, we only conducted experiments on ASPEC Japanese--English and JaRuNC Japanese--Russian for JASS (BLEU results on Japanese--Russian were excessively low for comparison).} Our results indicate that BMASS, BRSS, and HFSS significantly outperform the state-of-the-art MASS pre-training, whereas PMASS yields marginal improvements. Furthermore, we demonstrate that linguistically-driven multi-task pre-training methods (JASS \& ENSS) lead to further improvements of up to +2.9 BLEU points for Japanese to English, +2.7 BLEU points for English to Japanese, +4.3 BLEU points for Japanese to Chinese , +7.0 BLEU points for Chinese to Japanese, +0.5 BLEU points for English to Korean, and +1.3 BLEU points for Korean to English in low-resource scenarios, respectively.

Unlike in our previous study~\citet{mao-etal-2020-jass}, we provide substantial analyses for evaluating the translations generated by JASS and ENSS, which focus on the relationship between different pre-training tasks, and the specific adequacy and fluency of corresponding translations. Specifically, we validate the superior translation adequacy improvement of linguistically-driven methods by implementing automatic adequacy evaluation using LASER, human evaluation, and case study. To confirm the complementary nature between the masked language model and reordering the pre-training task, we performed an evaluation of the pre-training accuracy.

We expect this study to extend the usefulness of linguistically-driven pre-training methods for more low-resource language pairs and compensate for the defects of~\citet{mao-etal-2020-jass} in terms of the empirical evaluation. The contributions of this study can be summarized as follows. 

\begin{enumerate}
    \item \textbf{BMASS and BRSS:} Linguistically-driven novel pre-training methods for NMT involving Japanese.
    \item \textbf{PMASS and HFSS:} Linguistically-driven novel pre-training methods for NMT involving English (can be theoretically implemented on any SVO language).
    \item \textbf{Multi-task pre-training (JASS and ENSS)}: We demonstrate that multi-task training through the combination of the masked language model and reordering task (BMASS+BRSS \& MASS+HFSS) leads to better performance. Particularly, BMASS and BRSS can complement each other more if they are performed based on analogous syntactic units.
    \item \textbf{Empirical evaluation:} Comparisons among MASS, BART, JASS, ENSS and newly added baseline methods (MultiMASS and Deshuffling) for 6 translation directions and 3 different domains in several data size settings to identify situations in which each technique can be the most effective compared to other techniques.
    \item \textbf{Analyses:} Linguistic and statistical analyses of pre-training methods, their inter-relationships, and corresponding translations.
\end{enumerate}

\section{Related Work}

\subsection{Low-resource Neural Machine Translation}
There are mainly three lines of work related to improving NMT in low-resource situations: cross-lingual transfer, data augmentation, and monolingual pre-training. These approaches are potentially complementary. Our work belongs to the monolingual pre-training category.

Cross-lingual transfer addresses the low-resource issue by using data from different language pairs. One can use a richer language pair~\cite{zoph-etal-2016-transfer} or several language pairs simultaneously~\cite{dabre-etal-2019-exploiting,dong-etal-2015-multi}.~\citet{rudramurthy19} also proposed reordering the assisting languages to be similar to a low-resource language. 

Data augmentation involves the creation of synthetic bilingual data from monolingual data. In the popular back-translation approach~\cite{edunov-etal-2018-understanding,hoang-etal-2018-iterative,sennrich-etal-2016-improving}, the source side of the data is synthesized using an MT system to back-translate the target side data. Recently,~\citet{zhou-etal-2019-handling} proposed the creation of this source side through rule-based reordering via word-to-word translation.

In monolingual pre-training approaches, all or part of a model is first trained on tasks that require monolingual data.\footnote{This is an instance of ``transfer learning," similar to cross-lingual transfer. ``Pre-training" often implies that the training task differs from the target task.} Pre-training has enjoyed significant success in other NLP tasks with the development of GPT~\cite{Radford2018ImprovingLU}, BERT~\cite{devlin-etal-2019-bert}, and several others~\cite{Peters:2018, DBLP:conf/aaai/SunWLFTWW20, DBLP:conf/nips/YangDYCSL19}.

Pre-training schemes such as BERT were designed for natural language understanding (NLU) tasks and they are not directly suitable for NMT.~\citet{DBLP:conf/nips/ConneauL19} and~\citet{ren-etal-2019-explicit} proposed multilingual variants. However, they trained the encoder and decoder independently. To address this,~\citet{song2019mass} recently proposed MASS, a new state-of-the-art NMT pre-training task that jointly trains the encoder and decoder. Our approach develops on the initial idea of MASS, but adds more diverse and linguistically-motivated training objectives. 

Linguistic information is known to be useful for NMT~\cite{sennrich-haddow-2016-linguistic}, especially in low-resource scenarios. Outside of pre-training, studies~\cite{rudramurthy19,zhang-zong-2016-exploiting,zhou-etal-2019-handling} have successfully used a linguistically-motivated reordering similar to that of our BRSS task.~\citet{DBLP:conf/aaai/SunWLFTWW20} used linguistically-motivated pre-training tasks for text understanding. To the best of our knowledge, there are no studies on linguistically-motivated pre-training tasks for NMT.

\subsection{Pre-training Tasks for Neural Machine Translation}
\label{pre-tasks}
After the appearance of BERT~\cite{devlin-etal-2019-bert}, several pre-training methods have been proposed to enhance NMT~\cite{DBLP:conf/nips/ConneauL19,lewis-etal-2020-bart,lin-etal-2020-pre,DBLP:journals/tacl/LiuGGLEGLZ20,DBLP:journals/jmlr/RaffelSRLNMZLL20,ren-etal-2019-explicit,siddhant-etal-2020-leveraging,song-etal-2020-pre,song2019mass,wang-etal-2019-denoising,wang-etal-2020-multi-task,yang-etal-2020-csp}. Particularly, \citet{song2019mass} proposed a random span reconstruction task to pre-train a sequence-to-sequence framework for NMT; \citet{wang-etal-2019-denoising} first proposed using shuffling, deleting, and replacing operations to implement the denoising pre-training for the NMT system; thereafter, \citet{lewis-etal-2020-bart} combined the denoising methods with the masked language model pre-training of \citet{song2019mass}, and provided detailed empirical results for a large number of language pairs; mBART~\cite{DBLP:journals/tacl/LiuGGLEGLZ20} is a multilingual sequence-to-sequence denoising pre-training that is pre-trained through denoising tasks on 25 languages including Japanese, English, Chinese, Russian, and others, and it can be deemed as an extension of \citet{lewis-etal-2020-bart}; other studies focus on leveraging the cross-lingual supervision between languages through word alignment~\cite{lin-etal-2020-pre}, phrase alignment~\cite{ren-etal-2019-explicit}, sentence-level alignment~\cite{DBLP:conf/nips/ConneauL19}, code-switching technique~\cite{yang-etal-2020-csp}, or assisting languages (shared scripts)~\cite{song-etal-2020-pre}.

Among the above-mentioned pre-training techniques for NMT, we observe that no study has focused on leveraging specific linguistic features for NMT. Syntactic span-masking~\cite{zhou-etal-2020-limit} and semantic-aware BERT~\cite{DBLP:conf/aaai/0001WZLZZZ20} have been proposed using linguistically-driven supervision for language understanding tasks. However, linguistically-driven methods for sequence-to-sequence pre-training should be considered and explored. 

Studies have also focused on improving MASS. \citet{siddhant-etal-2020-leveraging} adapted MASS in multilingual scenarios; \citet{qi-etal-2020-prophetnet} proposed using an n-stream self-attention mechanism to enhance MASS for language generation tasks. No previous study has attempted to enhance MASS from a linguistic perspective, which will be explored in our study.

Moreover, \citet{wang-etal-2020-multi-task} highlighted that multitask learning can significantly benefit multilingual NMT. In addition to the MT task, the essential jointly-learned tasks should be masked langauge model task and denoising (reconstruction) task, which are two basic pre-training styles based on which we propose our linguistically-driven methods.

\section{Preliminary Backgrounds}
In this section, we introduce the preliminary backgrounds of pre-training and fine-tuning for NMT and MASS, which serve as the backbone for this study.

\subsection{Pre-training and Fine-tuning for NMT}

\begin{figure}[t]
\begin{center}
\includegraphics[width=0.85\linewidth]{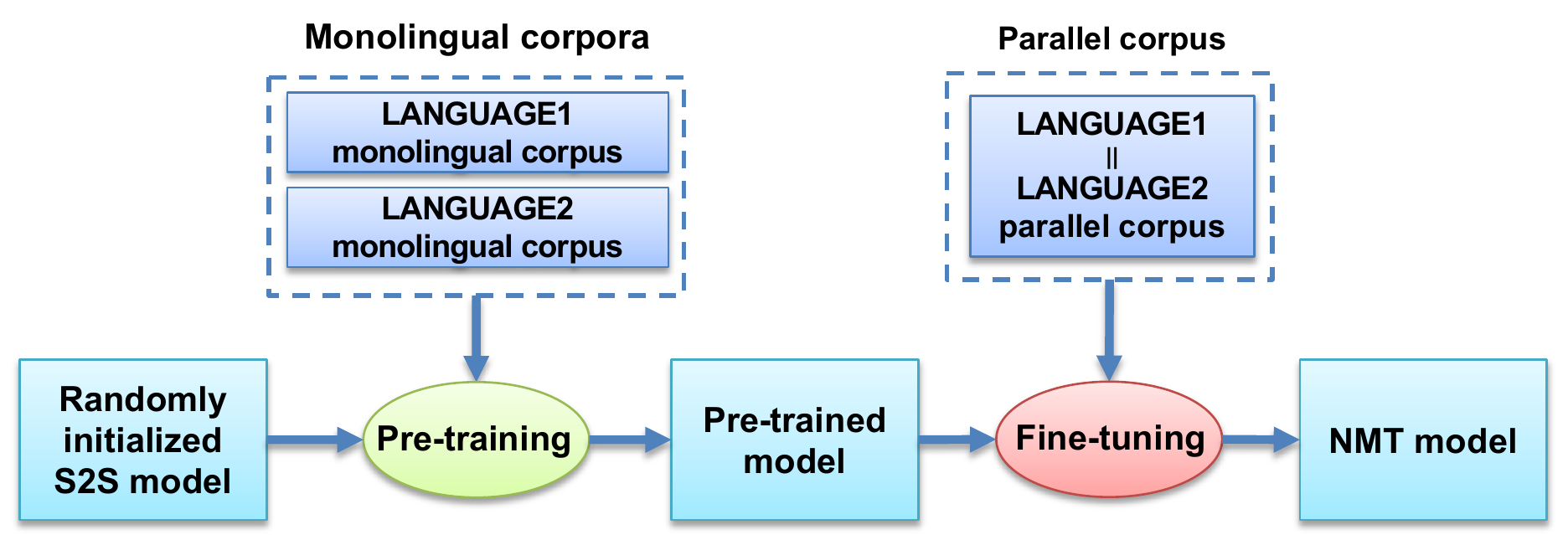} 
\caption{\textbf{Pre-training and fine-tuning for NMT.} ``S2S" denotes sequence-to-sequence.}
\label{pretraining}
\end{center}
\end{figure}

We first introduce the pre-training and fine-tuning pipelines for the NMT. As shown in Figure~\ref{pretraining} below, we first utilize monolingual corpora to pre-train the initialized sequence-to-sequence model. Subsequently, we use a parallel corpus of languages of interest to fine-tune the pre-trained models. The fine-tuned model was the final NMT model. All the experiments in this study will be conducted on the basis of this pre-training and fine-tuning pipeline for NMT.

\subsection{MASS}

\begin{figure}[t]
\begin{center}
\includegraphics[width=0.9\linewidth]{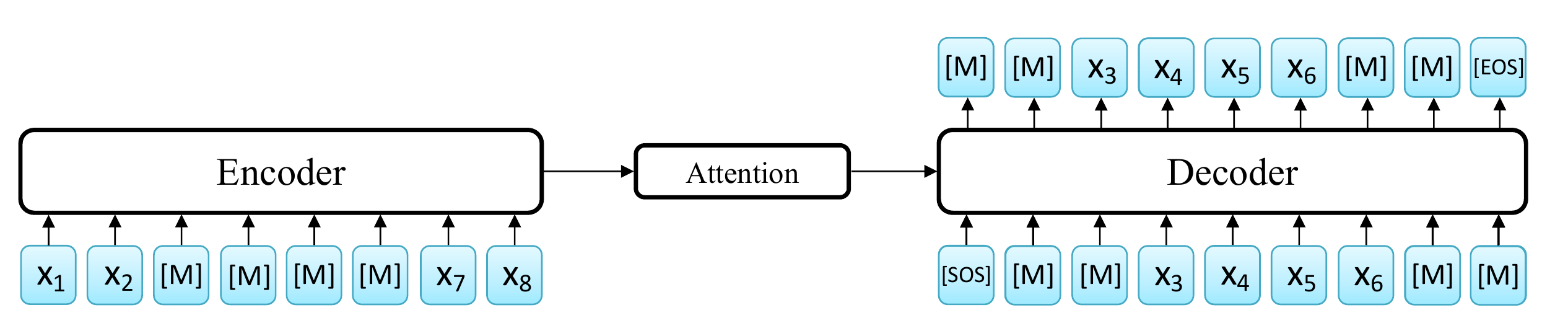} 
\caption{\textbf{Sequence-to-sequence structure for MASS.} $x_i$ represents a token and $x_3$ to $x_6$ are consecutive tokens to be masked/predicted.}
\label{fig.1}
\end{center}
\end{figure}

MASS is a pre-training method for NMT proposed by \citet{song2019mass}.
As shown in Figure~\ref{fig.1}, in MASS pre-training, the input is a sequence of tokens where a part of the sequence is masked and the output is a sequence where the masking is inverted. 

We consider $x\in \mathcal{X}$, which is a sequence of tokens where $\mathcal{X}$ is a monolingual corpus. Additionally, we consider the token span $C = [p_{i}, p_{j}]$, where $0< p_{i}\leq p_{j}\leq len(x)$ and $len(x)$ are the number of tokens in sentence $x$.
We denote the masked sequence by $x^{C}$, where tokens in positions from $p_{i}$ to $p_{j}$ in $x$ are replaced by a mask token $[M]$. $x^{!C}$ is the sequence with an inverted mask, that is, where tokens in positions other than the aforementioned fragments are replaced by the mask token $[M]$. 
In MASS, the pre-training objective is to predict the masked fragments in $x$ using an encoder-decoder model, where $x^{C}$ is the input to the encoder and $x^{!C}$ is the target output of the decoder. The log-likelihood objective function is
\vspace{-0.7cm}

\begin{eqnarray}
\label{eqn:mass}
\mathcal{L}_{mass}(\mathcal{X}) &=& \frac{1}{|\mathcal{X}|} \sum_{x \in \mathcal{X}} \log P\left(x^{!C} | x^{C}, \boldsymbol{\theta}\right).  
\end{eqnarray}
where $\boldsymbol{\theta}$ denotes the model parameters.
The number of tokens to be masked is a hyperparameter of the MASS. The NMT model is jointly pre-trained with the MASS task for both the source and target languages. 

\section{Proposed Methods} \label{sec:methoddescription}
In this section, we describe JASS and ENSS, which are our proposed pre-training techniques.

\subsection{Proposed Methods for Japanese}
\label{sec:jass}

Our methods are based on the ideas of the original MASS and are improved by jointly learning multiple linguistics-aware tasks. For Japanese, we propose a bunsetsu-based MASS (BMASS) pre-training and bunsetsu reordering-based sequence-to-sequence (BRSS) pre-training. Their combination, Japanese-specific sequence-to-sequence (JASS) pre-training, is introduced in the following section.

\subsubsection{Bunsetsu}

\begin{figure}[t]
\centering
\includegraphics[width=0.9\linewidth]{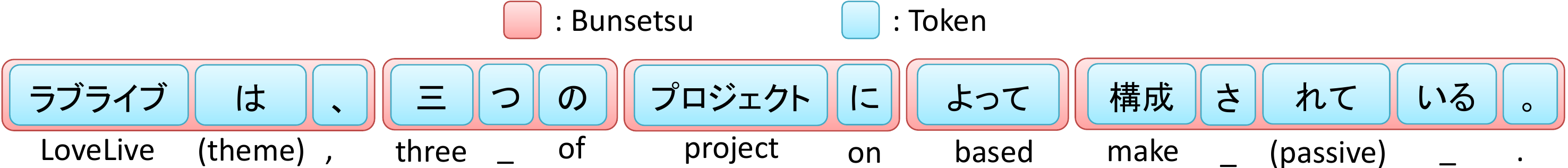}
\caption{Word and bunsetsu segmentations for a Japanese sentence with meaning ``LoveLive is made of three projects." In word for word English translations, ``\_" represents words with no meaningful translations.}
\label{fig:bunsetsu}
\end{figure}

Bunsetsu is the syntactic component of Japanese sentences~\cite{kurohashi--EtAl:1994,jumanpp2}. It is equivalent to the concepts of noun phrases or verb phrases in English syntax and it constitutes a minimal unit of meaning. The concept of ``word" is ambiguous for writing systems such as Japanese where word-separators are not applicable, and Japanese segmenters~\cite{kurohashi--EtAl:1994,jumanpp2} can segment Japanese sentences either in words or bunsetsus. Therefore, bunsetsu is also more likely to correspond to a well-defined entity or concept than words. Figure~\ref{fig:bunsetsu} illustrates the difference between the word- and bunsetsu-level segmentation. Each bunsetsu contains self-contained information and case markers, which indicate its relation with other bunsetsus. Based on the bunsetsu, we introduce our proposed pre-training techniques for the Japanese.

\subsubsection{BMASS} \label{sec:bmass}
We propose BMASS, which leverages syntactically parsed Japanese monolingual data for sequence-to-sequence pre-training. MASS pre-trains an NMT model by making it predict random parts of a sentence given their context, whereas BMASS involves making the model predict a set of bunsetsus given the contextual bunsetsus. We expect this will allow the model to learn about bunsetsus and thereby focus on predicting meaningful subsequences instead of random, albeit fluent subsequences.

\begin{figure}[t]
\begin{center}
\includegraphics[width=1\linewidth]{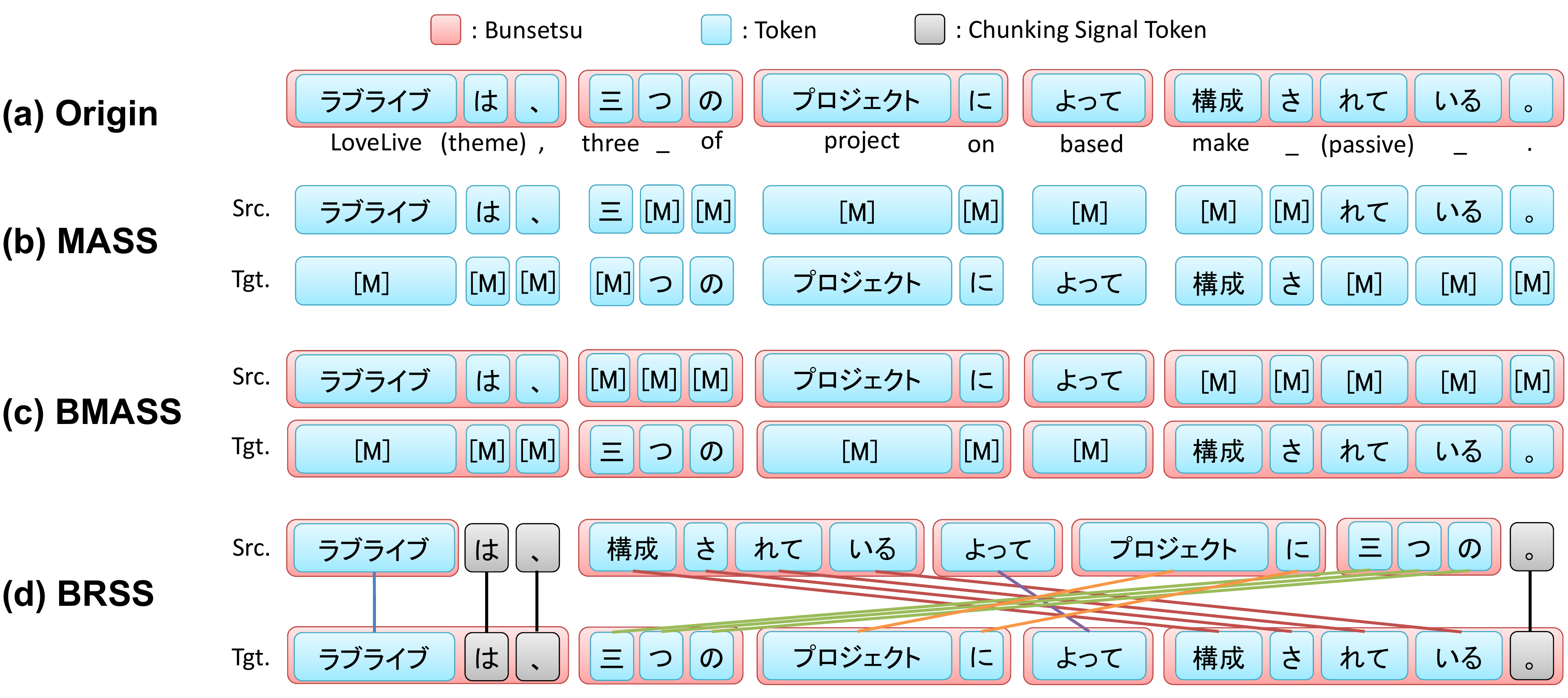} 
\caption{Example of source and target for MASS, BMASS, and BRSS with the meaning ``LoveLive is made of three projects."}
\label{fig.2}
\end{center}
\end{figure}

To perform BMASS, we modify the definition of mask $C$ in Equation~\ref{eqn:mass}: $C = [[p_{i_1},p_{j_1}],[p_{i_2},p_{j_2}],...$ $[p_{i_n},p_{j_n}]]$, where $0< p_{i_1}\leq p_{j_1}\leq p_{i_2}\leq p_{j_2}\leq ... p_{i_n}\leq p_{j_n}\leq len(x)$. 
Term $len(x)$ denotes the number of tokens in sentence $x$. Subsequently, the $k-th$ position span from $p_{i_k}$ to $p_{j_k}$ corresponds to the start and end of a specific bunsetsu in a Japanese sentence. Consequently, we denote the BMASS loss as $\mathcal{L}_{bmass}$. The main difference between MASS and BMASS is that in MASS, we mask random token spans, whereas in BMASS, we only mask tokens spans that are complete bunsetsus. The number of bunsetsus to be masked constitutes a hyperparameter for BMASS. Figures~\ref{fig.2}-b and~\ref{fig.2}-c provide training pairs for MASS and BMASS.

Note that our BMASS pre-training task differs from the entity masking task of ERNIE~\cite{DBLP:conf/aaai/SunWLFTWW20} and random span masking of SpanBERT~\cite{joshi-etal-2020-spanbert}. ERNIE and SpanBERT have been proposed without using syntactic units and they are employed in natural language understanding downstream tasks.

\subsubsection{BRSS} \label{sec:rss}
Japanese sentences are typically in an SOV word order that can be reordered to SVO to reduce the difficulty of translation to languages with SVO order. We first define a simple process for reordering a (typically SOV) Japanese sentence into a ``SVO Japanese" pseudo-sentence that will be used in BRSS.
There are several previous studies on reordering a SOV-ordered sentence to a SVO-ordered sentence \cite{hoshino-etal-2013-two,komachi2006phrase}. In our case, to consistently leverage bunsetsu units in Japanese with BMASS, we propose bunsetsu-based reordering, which is able to generate an SVO-ordered Japanese sentence while retaining syntactic information at the bunsetsu-level. We first define ``chunking signal words" as any punctuation mark or the topic marker ``\begin{CJK}{UTF8}{ipxm}は\end{CJK}." The reordering process is as follows:
\begin{enumerate}
    \item split the sentence into bunsetsus
    \item select sequences of bunsetsus bounded by chunking signal words
    \item simply reverse the order of the bunsetsus in these sequences without using rules
\end{enumerate}

We can now propose BRSS, which involves a Japanese sentence and its reordered version obtained using the aforementioned procedure. Refer to Figure~\ref{fig.2}-d as an example of a bunsetsu-reordered sentence. The pre-training objective was a reordering task. We expect that this will allow the system to learn the structure of the Japanese language, and prepare it for the reordering operation it will have to perform when translating to a language with different grammar. Although BRSS task is constructed by simple rules, the predictions for the bunsetsu boundaries and orders are expected to equip the model with abundant linguistic knowledge. We have two choices from which we can make the NMT system predict the original sentence given the reordered sentence (BRSS.F) or vice-versa (BRSS.R). We will experiment with both options.

\subsection{Proposed Methods for English}
Similar to the proposed methods for Japanese, we propose two linguistically-driven methods for English that are based on the MASS language model and reordering sequence-to-sequence language model, respectively. One is phrase structure-based MASS (PMASS), and the other method is head finalization-based sequence-to-sequence pre-training (HFSS). The combination of PMASS, HFSS, and ENSS is introduced in the next section. Before introducing our proposed methods for English, we first provide background information on head-driven phrase structure grammar and head finalization, which forms our linguistically-driven methods.

\subsubsection{Head-driven Phrase Structure Grammar}
As opposed to dependency-based grammar, head-driven phrase structure grammar (HPSG) \cite{10.5555/42293,PollardSag94} is lexicalism-based grammar that focuses on generalizing phrase structures. HPSG primarily handles word and phrase signs in a sentence in terms of their syntactic and semantic roles. Thus, HPSG should be an appropriate parsing rule for extracting phrase structures in sentences and applying the following proposed pre-training techniques. Figure~\ref{hf} (left) shows an instance of parsing an English sentence using HPSG grammar.

\subsubsection{Head Finalization}
\label{bg.hf}

\begin{figure}[t]
\begin{center}
\includegraphics[width=0.75\linewidth]{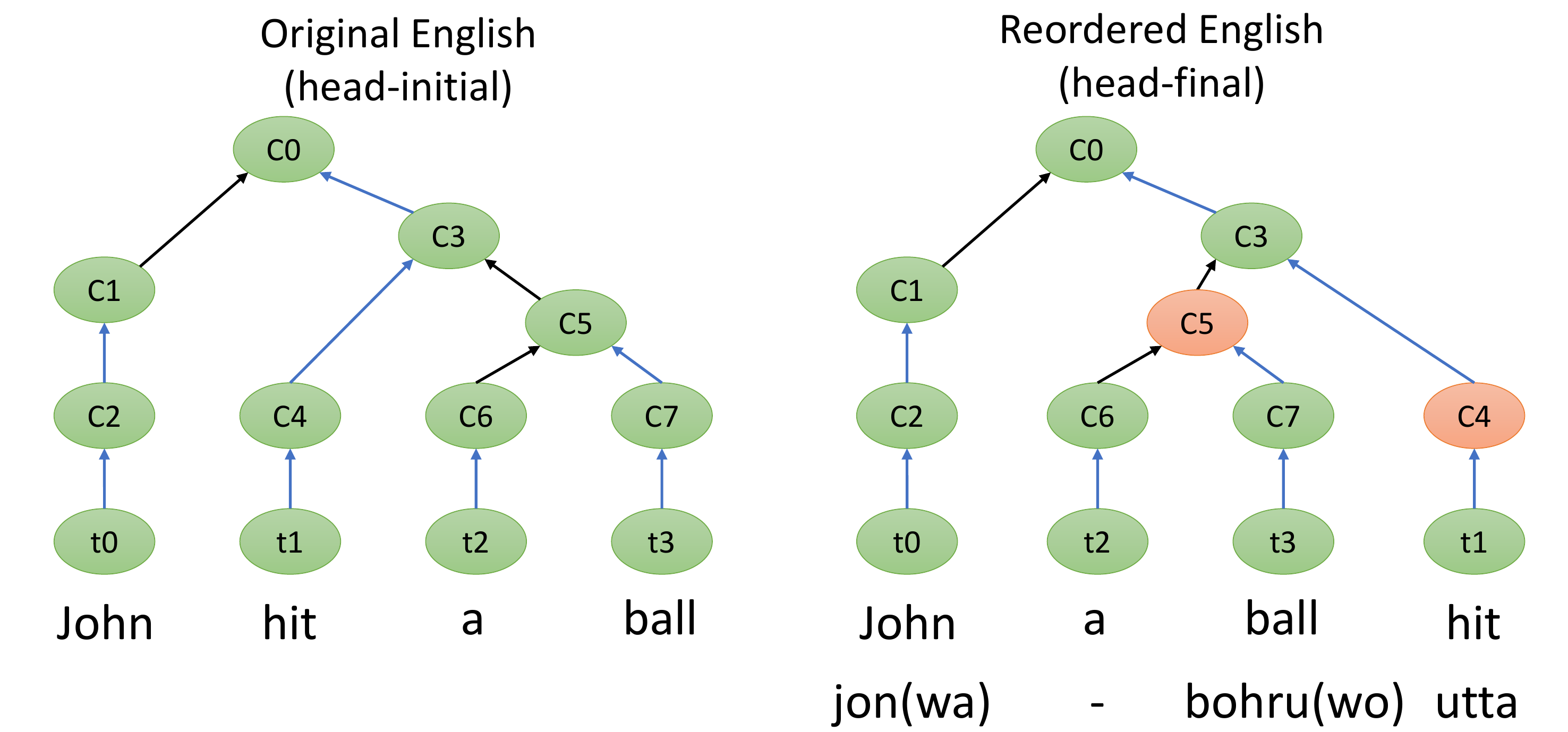} 
\caption{Example of HPSG parsing result and head finalization. Head finalization reorders an English sentence into a Japanese-like sentence.~\cite{isozaki-etal-2010-head} Blue arrows denote the ``head. ”}
\label{hf}
\end{center}
\end{figure}

Using the above-mentioned HPSG, sentences in any language can be characterized using phrase structures. From the definition of a phrase, the ``head" of a phrase is subsequently defined as the syntactically determinant part in a phrase. In other words, ``head" determines the syntactic category of the phrase and its ``dependents." Particularly, English is referred to as a ``head-initial" language because ``head" appears before its ``dependents," whereas Japanese is referred to as a "head-final" language because ``head" usually appears after ``dependents" in a phrase.

The deliberate phrase structures provided by the HPSG parser are utilized in several scenarios in the NLP. Particularly, \citet{isozaki-etal-2010-head} proposed a simple reordering rule for the SVO language (head-initial languages) by using the phrase structure information provided by the HPSG parser. Figure~\ref{hf} shows an example of reordering an English sentence to be an SOV-like sentence on the basis of the result of HPSG parsing. By reordering sentences in SVO languages such as English to be SOV-like sentences, the performance of statistical machine translation (SMT) is improved. Particularly, \citet{isozaki-etal-2010-head} first proposed head finalization and applied it to English-to-Japanese SMT; \citet{han-etal-2012-head} applied it to Chinese-to-Japanese SMT and obtained significant improvements; more recently, \citet{zhou-etal-2019-handling} utilized this reordering technique to generate synthetic parallel sentences in the back-translation phase when translating SOV and SVO languages. In this study, we utilize this reordering rule in the pre-training phase for NMT (see Section~\ref{hfss}).

\subsubsection{PMASS}

\begin{figure}[t]
\begin{center}
\includegraphics[width=0.8\linewidth]{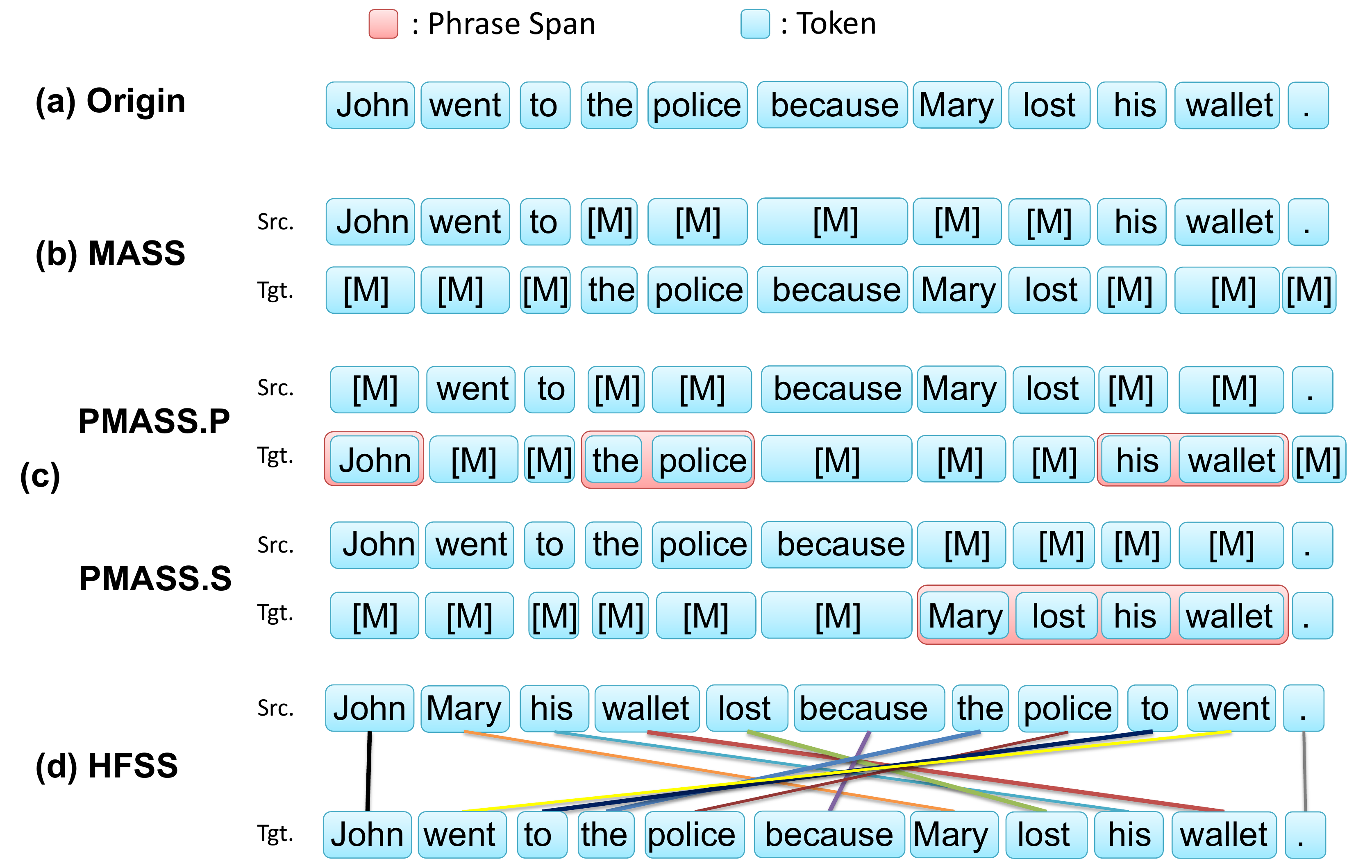}
\caption{Example of source and target for MASS, PMASS, and HFSS of a sentence in English.}
\label{pmass+hfss}
\end{center}
\end{figure}

We propose PMASS by leveraging phrase-span information in an English sentence. In general, we perform PMASS pre-training by limiting the masked tokens in MASS to be an entire phrase span. Thus, for masking plural phrase spans, we denote it as PMASS.P. For masking only a single phrase span, we denote it as PMASS.S. Particularly, the source and target for PMASS.P and PMASS.S pre-training can be generated using our proposed phrase-masking algorithms described in Appendix~\ref{algo}. Inspired by MASS, we force the number of masked tokens to be approximately half of the length of the sentence to guarantee the effectiveness of the sequence-to-sequence masked language model. Examples of PMASS.P and PMASS.S are presented in Figure~\ref{pmass+hfss}-c. We observe that several phrase spans in PMASS.P and a single long phrase span in PMASS.S are masked. We expect such special masking patterns to force the NMT system to extract more phrase-level syntactic information in the pre-training phase.

\subsubsection{HFSS}
\label{hfss}
We propose HFSS using the head finalization technique~\cite{isozaki-etal-2010-head} for pre-training English. As shown in Figure~\ref{pmass+hfss}-d, the pre-training task is also a reordering task that simulates the translation from SOV languages to English. More precisely, the source sentence for sequence-to-sequence pre-training is the reordered (SOV-like or head-finalized) English sentence, and the target sentence is the original English monolingual sentence. We expect HFSS to help the system learn the word reordering pattern of the translation between head-initial (SVO) and head-final (SOV) languages in advance.

According to the prior experiments for Japanese (see~\citet{mao-etal-2020-jass} and~\ref{nmtresults}), BRSS.F consistently outperforms BRSS.R. In addition, BART~\cite{lewis-etal-2020-bart} also claims that reconstructing the original sentence benefits the language generation tasks. Therefore, we do not distinguish HFSS with HFSS.F and HFSS.R (HFSS.F performs pre-training with the SOV-SVO pattern, whereas HFSS.R performs the reverse pattern).\footnote{More precisely, HFSS.F denotes the source sentence of the head-finalized English sentence and the target sentence of the original English sentence. HFSS.R indicates the source sentence of the original English sentence and the target sentence of the head-finalized English sentence.} Instead, we directly defined HFSS using the pre-training pattern of HFSS.F. Moreover, HFSS is performed on the basis of head finalization, which utilizes the results from HPSG parsers. This is consistent with PMASS in which we extract phrases using HPSG-parsing results.

We develop our proposal on English through head finalization, whereas for SOV languages such as Japanese, it is unmanageable to reorder SOV sentences to SVO-like sentences~\cite{isozaki-etal-2010-head}. Furthermore, HFSS can be used for all head-initial languages apart from English, as well-developed reordering rules have been proposed and demonstrated to be effective for NMT. However, BRSS can only be implemented for Japanese-involved translation pairs because bunsetsu information is required to establish the source and target sentences for sequence-to-sequence pre-training.

\subsection{Multi-task Pre-training}
Multi-task pre-training objectives lead to a robust initial state for NMT systems~\cite{lewis-etal-2020-bart,DBLP:journals/jmlr/RaffelSRLNMZLL20}. Because our proposed methods can also be categorized into two groups of pre-training tasks, we propose a multi-task pre-training task for both Japanese and English. 

We define JASS pre-training, which is a combination of the two previous procedures: BMASS and BRSS. Our actual pre-training will consist of the joint execution of these two pre-training sessions. Therefore, the pre-training objective for JASS is 
\begin{eqnarray}
\mathcal{L}_{jass}(\mathcal{X}_{ja}) = \mathcal{L}_{bmass}(\mathcal{X}_{ja}) + \mathcal{L}_{brss}(\mathcal{X}_{ja})
\end{eqnarray}
where $\mathcal{X}_{ja}$ represents the monolingual corpus of Japanese, and $\mathcal{L}_{brss}$ denotes the reordering loss using the forward or reverse variants mentioned in Section~\ref{sec:rss}. We expect BMASS \& BRSS to jointly learn syntactic knowledge and BRSS to learn word ordering knowledge.

For English, we similarly define ENSS pre-training, which combines PMASS and HFSS. More precisely, the training objective is:
\begin{eqnarray}
\mathcal{L}_{enss}(\mathcal{X}_{en}) = \mathcal{L}_{pmass}(\mathcal{X}_{en}) + \mathcal{L}_{hfss}(\mathcal{X}_{en})
\end{eqnarray}
where $\mathcal{X}_{en}$ denotes the monolingual corpus of English, $\mathcal{L}_{pmass}$ the PMASS.P or PMASS.S loss, and $\mathcal{L}_{hfss}$ the reordering loss of HFSS.

JASS is specifically designed for Japanese, whereas theoretically, ENSS can be transplanted onto any SVO language as long as we can extract the phrase structure information of the corresponding language from a HPSG parser.

We also mixed JASS pre-training for Japanese with MASS pre-training for the other languages involved in the translation. In practice, we therefore designated using JASS pre-training for Japanese monolingual data with BMASS and BRSS objectives, along with ``other languages" monolingual data with the MASS objective. Similarly, for English, ENSS pre-training consists of PMASS \& HFSS for English and MASS for ``other languages" involved in fine-tuning translation pair.

We also consider attempting the combination of our proposed linguistically-driven methods with a strong baseline pre-training objective, MASS, which we refer to as MASS + JASS (or ENSS) in the subsequent sections. To allow the pre-training model to determine the language and sub-task (MASS, BMASS, BRSS, PMASS, and HFSS) that it should perform, we prepend tags to inputs similar to those used in~\citet{johnson-etal-2017-googles} (see Section~\ref{sec:preproc} for details).

\section{Experimental Settings}
In this section, we evaluate our pre-training methods on simulated low-resource scenarios for ASPEC Japanese--English~\cite{nakazawa-etal-2016-aspec}, Japanese--Chinese translations~\cite{nakazawa-etal-2015-overview}, and realistic low-resource scenarios for Wikipedia Japanese--Chinese~\cite{chu-etal-2014-constructing,DBLP:journals/talip/ChuNK16} and News English--Korean~\cite{park-etal-2016-korean} translations.

\subsection{Datasets}
We used monolingual data for pre-training and parallel data for fine-tuning. Refer to Table~\ref{data} for an overview. 

\begin{table}[h!]
\begin{center}
\begin{tabular}{cc c r}

      \toprule
      & Language& Dataset&Size\\
       \midrule
      \multirow{3}{*}{Monolingual}&Ja& Common Crawl& 22M\\
      &Zh& Common Crawl & 22M\\
      &En& Common Crawl & 22M\\
      &Ko& Common Crawl & 22M\\
       \midrule
      \multirow{4}{*}{Parallel}&Ja-En& ASPEC-JE&1M \\
     &Ja-Zh&ASPEC-JC&670k\\
     &Ja-Zh&Wikipedia&258k\\
     &En-Ko&News&94k\\
      \bottomrule
\end{tabular}
\caption{Overview of training data. ``Size" denotes the number of the monolingual sentences or parallel sentences.} 
\label{data}
 \end{center}
\end{table}

\noindent\textbf{Monolingual data:} For pre-training, we use monolingual data of 22M lines each for Japanese, English, Chinese, and Korean, randomly sub-sampled from Common Crawl mentioned in the official WMT monolingual training data.\footnote{\url{http://www.statmt.org/wmt19/translation-task.html}} \footnote{Different from~\citet{mao-etal-2020-jass}. Currently, we unify the monolingual corpus domains for all the languages for fairer comparisons.} For pre-training in Japanese--English and English--Korean, given that these two languages have different scripts and thus have few common words, the pre-training objectives for each language will work separately, even though they are performed jointly for two languages. However, for pre-training in Japanese and Chinese, they share more characters, which indicates that the monolingual pre-training tasks will be run in a pseudo-cross-lingual manner. Thus, we also expect to see whether such pre-training will benefit from more fine-tuning.

\noindent\textbf{Parallel Data:} We use scientific abstracts domain ASPEC parallel corpus for training Japanese--English and Japanese--Chinese models. For Japanese--Chinese fine-tuning, we also utilize the Wikipedia parallel corpus, which is a real low-resource scenario. We use News parallel corpus for English--Korean, which is a low-resource dataset.

For ASPEC, we used the official training, development, and test splits provided by WAT 2019.\footnote{\url{http://lotus.kuee.kyoto-u.ac.jp/WAT/WAT2019/index.html\#task.html}} \footnote{For ASPEC Japanese--English, we use the first 1M parallel sentences. Parallel sentences for different fine-tuning size settings were randomly sampled from the selected 1M dataset.} For Wikipedia, we used the dataset released by Kyoto University.\footnote{\url{http://nlp.ist.i.kyoto-u.ac.jp/EN/index.php?Wikipedia\%20Chinese-Japanese\%20Parallel\%20Corpus}} For News, we use dataset provided by \citet{park-etal-2016-korean}.\footnote{\url{https://sites.google.com/site/koreanparalleldata}}

\subsection{Pre-processing} \label{sec:preproc}
We tokenize the monolingual data by using the Moses tokenizer for English and Korean,\footnote{\url{https://github.com/moses-smt/mosesdecoder}} Jumanpp for Japanese,\footnote{\url{https://github.com/ku-nlp/jumanpp}} and jieba for Chinese.\footnote{\url{https://github.com/fxsjy/jieba}} We obtain the bunsetsu information by using KNP\footnote{\url{https://github.com/ku-nlp/pyknp}} and obtain the HPSG parsing results using enju.\footnote{\url{https://mynlp.is.s.u-tokyo.ac.jp/enju/}} Sentences with more than 175 tokens were removed. For each language pair, we constructed a joint vocabulary with 60 000 sub-word units through byte-pair encoding (BPE)~\cite{sennrich-etal-2016-neural} on the concatenated monolingual corpora involved during pre-training.\footnote{Particularly, 30 000 BPE merging operations will lead to a joint vocabulary with a size of approximately 60 000 for Japanese--Chinese and English-Korean}, whereas 40 000 BPE merge operations is set for Japanese-English. In the multi-task pre-training, each sentence is prepended with a task token $[MASS]$, $[BMASS]$, $[BRSS]$, $[PMASS]$, or $[HFSS]$, and a language token $[Ja]$, $[En]$, $[Zh]$, or $[Ko]$.\footnote{As an implementation trick, we recommend to unify the task tag for the same group of tasks, e.g. use the same tag for $[BRSS]$ and $[HFSS]$.} This ensures that the model learns to distinguish between different pre-training objectives and languages. This token can be used when monolingual pre-training is conducted jointly by multiple languages and multiple tasks.

\subsection{Training and Evaluation Details}
In our experiments, we used the open-source OpenNMT~\cite{klein-etal-2017-opennmt} implementation of the transformer~\cite{NIPS2017_7181} NMT model.\footnote{\url{https://github.com/OpenNMT/OpenNMT-py}} The hyperparameters are set to the transformer-big setting in OpenNMT. Particularly, our model has a 6-layer encoder and decoder, a hidden size of 1024, feed-forward hidden layer size of 4096, batch size of 4096, dropout rate of 0.3, and 16 attention heads. An ADAM optimizer with a learning rate of $10^{-4}$ was used for both pre-training and fine-tuning. All the pre-training tasks are run until convergence on four TITAN V100 GPU cards occurs, and fine-tuning uses only one GPU. It took approximately two days for each pre-training run. Mixed precision training~\cite{DBLP:conf/iclr/MicikeviciusNAD18} was used for both pre-training and fine-tuning. For multi-task pre-training, data are randomly shuffled such that even in each mini-batch, different pre-training objectives appear, corresponding to a real joint pre-training. Our proposed pre-training methods converge within the similar training time as compared to that of MASS.

Pre-training tasks are evaluated using perplexity, and the checkpoint with the lowest pre-training perplexity was selected for fine-tuning. We used BLEU~\cite{papineni-etal-2002-bleu} for automatic evaluation, adequacy, and fluency for human evaluation. We performed early stopping using 1-gram accuracy and perplexity on the development set. We evaluated the statistical significance of our BLEU scores through bootstrap resampling~\cite{koehn-2004-statistical}.

\subsection{Baselines}
In addition to MASS, we employ the ``text infilling" in BART as another main baseline.\footnote{Text infilling has been demonstrated as the most effective pre-training objective for NMT among several objectives in BART~\cite{lewis-etal-2020-bart}.} We also define two pre-training baselines for comparison with our proposed methods. They are named multi-span-based MASS (MultiMASS) and deshuffling. Moreover, the joint training with MASS and deshuffling was set as the multi-task pre-training baseline. All of the baselines are as follows:

\noindent{\textbf{Baselines without pre-training.}} First, we employ the vanilla transformer big as the baseline without pre-training because all of the pre-training methods are based on this model structure. Moreover, following~\citet{araabi-monz-2020-optimizing}, we also present the best performance for low-resource NMT by using Transformer model. Hyperparameter details are shown in Appendix~\ref{appa}.

\noindent{\textbf{MASS.}} Using the same settings as in~\citet{song2019mass}.

\noindent{\textbf{BART (text infilling).}} Different from MASS, BART (text infilling) masks several token spans within a sentence by a single $[M]$ where span lengths are samples from Poisson distribution and the model is also required to predict the lengths of the masked spans. We use the same settings as in~\citet{lewis-etal-2020-bart}.\footnote{In order to conduct fair comparisons for our proposed methods, we only present the most effective sub-task, text infilling, within BART. The combination of text-infilling and sentence permutation is proven to be the best practice of BART. With regard to sentence permutation, we do not consider it in this paper because it is mainly designed for document NMT. When it comes to multi-sentence pre-training, sentence permutation and other possible patterns of multi-sentence linguistically-driven pre-training tasks should be explored and compared in future work.}

\noindent{\textbf{MultiMASS.}} MultiMASS is a baseline method added to help demonstrate the effectiveness of masking specific syntactic units such as bunsetsu or phrase spans in a sentence that we propose as BMASS and PMASS. 

\begin{figure}[t]
\begin{center}
\includegraphics[width=0.9\linewidth]{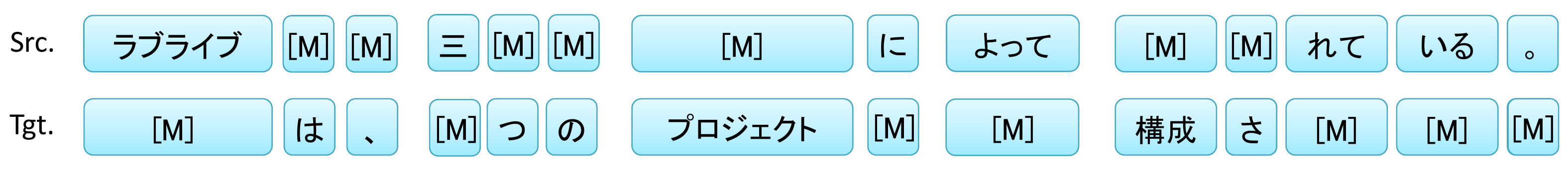} 
\caption{Example of source and target for MultiMASS with the meaning ``LoveLive is made of three projects."}
\label{MultiMASS}
\end{center}
\end{figure}

As shown in Figure~\ref{MultiMASS}, MultiMASS predicts several randomly masked tokens in a sentence, which differs from the single masked span in MASS, masked busetsu spans in BMASS, several phrase spans in PMASS.P, and a single phrase span in PMASS.S.

\begin{figure}[t]
\begin{center}
\includegraphics[width=0.9\linewidth]{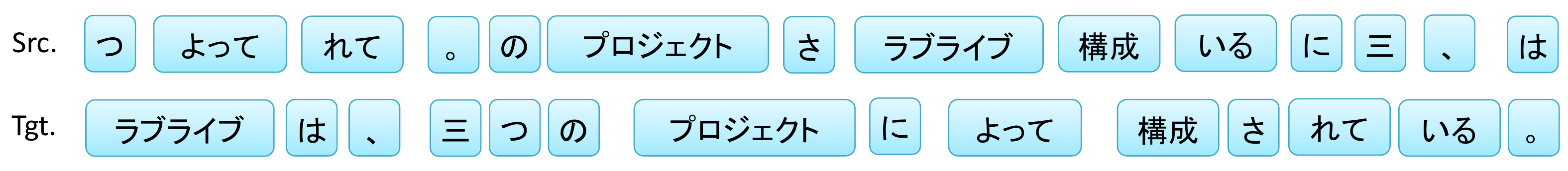} 
\caption{Example of source and target for deshuffling with the meaning ``LoveLive is made of three projects."}
\label{Deshuffling}
\end{center}
\end{figure}

\noindent{\textbf{Deshuffling.}} Deshuffling denotes the pre-training task of random shuffling-based sentence reconstruction, which is also a crucial pre-training task. We perform this pre-training task as another baseline to confirm the effectiveness of reordering syntactic units in BRSS and the reordering driven by head finalization of HFSS. A pre-training example is presented in Figure~\ref{Deshuffling}.

\noindent{\textbf{Multi-task Baseline.}}
The multi-task baseline is the combination of the respective best baseline methods from the masked language model and reordering pre-training. Thus, the multi-task baseline consists of MASS,\footnote{MASS outperforms MultiMASS, we therefore use MASS rather than MultiMASS. (See~\ref{nmtresults})} and deshuffling. The baseline is formulated as follows:
\begin{eqnarray}
\mathcal{L}(\mathcal{X}) = \mathcal{L}_{mass}(\mathcal{X}) + \mathcal{L}_{deshuffling}(\mathcal{X})
\end{eqnarray}
where $\mathcal{X}$ represents the monolingual corpora.

\subsection{Pre-trained Models}\label{sec:pretrainedmodels}

\begin{table}[h!]
\begin{center}
\resizebox{0.85\textwidth}{!}{
\begin{tabular}{cc c}

      \toprule
     \# & Pre-trained Model & Details\\
      \midrule
      \multicolumn{3}{l}{\textit{Main baseline}}\\
      \hline
      1 &MASS & Using the same settings as in \citet{song2019mass}.\\
      1* & BART (text infilling) & Using the same settings as in \citet{lewis-etal-2020-bart}.\\
      \midrule
      \midrule
      \multicolumn{3}{l}{\textit{Proposed methods for Japanese}}\\
      \hline
      2 &BMASS&	\begin{tabular}{c}
           Similar to MASS, we mask half the number of \\ bunsetsus during pre-training.
      \end{tabular}\\
\hline
      3 &BRSS& 
      \begin{tabular}{c}
        We separately pre-trained on the SVO--SOV (BRSS.F) \\ as well as SOV--SVO (BRSS.R) models.
      \end{tabular}
\\
\hline
      4 &JASS& Multi-task training of BMASS and BRSS.
\\
      \midrule
      \midrule
      \multicolumn{3}{l}{\textit{Combinations of proposed methods with MASS}}\\
      \hline
      5 & MASS+BMASS &	Multi-task training of MASS and BMASS.
\\
\hline
      6 & MASS+BRSS &Multi-task training of MASS and BRSS.\\
      \hline
      7 & MASS+BMASS+BRSS &Multi-task training of BMASS, BRSS and MASS.	\\
      \midrule
      \midrule
      \multicolumn{3}{l}{\textit{Other baselines for Japanese}}\\
      \hline
      8&MultiMASS (Ja) &
      \begin{tabular}{c}
           Based on MASS pre-training, several random tokens are \\ masked rather than one consecutive span.
      \end{tabular}
\\
\hline
      9&Deshuffling (Ja) &Random shuffling-based original sentence reconstruction.\\
      \hline
      10&MASS+Deshuffling (Ja)& Multi-task pre-training baseline for Japanese.\\
      \midrule
      \midrule
      \multicolumn{3}{l}{\textit{Proposed methods for English}}\\
      \hline
      11 &PMASS&	\begin{tabular}{c}
           Similar to MASS, we mask an entire phrase span based on the \\ head-driven phrase structure grammar. We \\ performed the experiments for PMASS.P and PMASS.S, respectively.
      \end{tabular}\\
\hline
      12 &HFSS& 
      \begin{tabular}{c}
        We train SOV (head finalized)—SVO (original) models for English.
      \end{tabular}
\\
\hline
      13 &ENSS& Multi-task training of MASS and HFSS.
\\
      \midrule
      \midrule
      \multicolumn{3}{l}{\textit{Other baselines for English}}\\
      \hline
      14&MultiMASS (En) &
      \begin{tabular}{c}
           Based on the MASS, several random tokens are \\ masked rather than one consecutive span.
      \end{tabular}
\\
\hline
      15&Deshuffling (En) &Random shuffling-based original sentence reconstruction.\\
      \hline
      16&MASS+Deshuffling (En) & Multi-task pre-training baseline for English.\\
      \midrule
      \midrule
      \multicolumn{3}{l}{\textit{Combination of the proposed methods for English and Japanese}}\\
      \hline
      17 & JASS+ENSS & Multi-task training of JASS and ENSS.\\
      \midrule
      \multicolumn{3}{l}{\textit{Baseline for \#17}}\\
      \hline
      18 & MASS+Deshuffling & Multi-task pre-training baseline for JASS+ENSS.\\
      \bottomrule
\end{tabular}
}
\caption{Settings of pre-trained models.} 
\label{pretrainedmodels}
 \end{center}
\end{table}

We pre-trained our NMT models by leveraging the monolingual data of the source and target languages. For Japanese, we can use MASS, BMASS, or BRSS, whereas for English, we can use MASS, PMASS, or HFSS. For Chinese and Korean, we use only the MASS. Particularly, we pre-trained different types of models in Table~\ref{pretrainedmodels}. Note that we use MASS for ENSS because PMASS underperforms MASS by a significant margin (see~\ref{nmtresults}).

\subsection{Fine-tuned NMT Models}
We fine-tuned to improve Japanese-English, English-Japanese, Japanese-Chinese, Chinese-Japanese, English--Korean and Korean--English translations. We trained the following NMT models:

\begin{enumerate}
    \item \textbf{Ja--En and En--Ja:} Japanese to English and English to Japanese models using from 3k to 50k parallel sentences randomly sampled from \textbf{ASPEC} for fine-tuning.
    \item \textbf{Ja--Zh and Zh--Ja:} Japanese to Chinese and Chinese to Japanese models using from 3k to 50k parallel sentences randomly sampled from \textbf{ASPEC} and \textbf{Wikipedia}, respectively, for fine-tuning.
    \item \textbf{En-Ko and Ko-En:} English to Korean and Korean to English models using 20k (randomly sampled) and 94k (full dataset) parallel sentences from \textbf{News} for fine-tuning.
\end{enumerate}

We compared these models with pre-trained model baselines and vanilla baselines, which are fully-supervised NMT models on the same data settings, but without pre-training. In addition, fine-tuning results under the high-resource scenarios (with more than 50k parallel sentences) are provided and discussed in~\ref{appendix:high-resource}.

\section{Results and Analyses}
Tables~\ref{ja-en},~\ref{ja-zh},~\ref{ja-zh2}, and~\ref{en-ko} contain the NMT BLEU results of our proposed methods for Japanese--English, Japanese--Chinese and English--Korean translation on various translation domains, respectively. Subsequently, we provide in-depth analysis for translation quality in terms of adequacy by using LASER~\cite{artetxe-schwenk-2019-massively}, human evaluation scores, specific cases for the real low-resource scenario of Wikipedia Ja-Zh. Finally, we conduct an investigation on the pre-training accuracy to analyze the difference between the pre-trained models and their complementation of each other, and present the results in middle/high-resource scenarios.

\subsection{NMT Results}
\label{nmtresults}

\begin{table}[hbtp]
\begin{center}
\resizebox{0.9\textwidth}{!}{
\begin{tabular}{lccccccccc}

      \toprule
     \multirow{2}{*}{\#} &\multirow{2}{*}{Model} &\multicolumn{4}{c}{Ja-En}&\multicolumn{4}{c}{En-Ja}\\
      %\midrule
      &&  3k &  10k & 20k & 50k & 3k & 10k & 20k & 50k\\
      \midrule
      \multicolumn{10}{l}{\textit{Main baselines}}\\
      0 & w/o pre, vanilla &0.8&2.1&3.5&16.1&1.1&2.7&5.1&19.4\\
      0* & w/o pre, optimized & 2.2 & 6.8 & 10.7 & 19.8 & 3.3 & 6.5 & 13.6 & 23.7 \\
      1 & MASS &8.8&13.8&17.2&21.2&9.1&16.0&20.6&25.0\\
      1* & BART (text infilling)  & 3.1  & 11.1 & 15.5 & 20.7 & 5.6  & 14.9 & 19.8 & 25.6$^\dag$ \\
      \midrule
      \midrule
      \multicolumn{10}{l}{\textit{Proposed methods for Japanese}}\\
      2 & BMASS &8.9&13.9&17.4&21.8&8.7&15.9&20.1&25.4\\
      3 & BRSS &8.8&14.9$^\dag$&18.1$^\dag$&22.0$^\dag$&10.0$^\dag$&17.3$^\dag$&21.0&26.0$^\dag$\\
      3 (R) & BRSS.R &8.2&14.3$^\dag$&17.7$^\dag$&21.7$^\dag$&10.0$^\dag$&17.2$^\dag$&20.5&25.7$^\dag$\\
      4 & JASS &10.6$^\dag$&15.7$^\dag$&\textbf{18.9}$^\dag$&\textbf{22.3}$^\dag$&\textbf{11.5}$^\dag$&17.7$^\dag$&21.6$^\dag$&26.5$^\dag$\\
      \hline
      \multicolumn{10}{l}{\textit{Combinations of proposed methods with MASS}}\\
      5 & 1 + 2 &9.2&14.8$^\dag$&17.7$^\dag$&21.7$^\dag$&9.7$^\dag$&16.6$^\dag$&20.9&25.9$^\dag$\\
      6 & 1 + 3 &\textbf{10.9}$^\dag$&15.9$^\dag$&18.3$^\dag$&22.2$^\dag$&11.0$^\dag$&17.7$^\dag$&\textbf{21.7}$^\dag$&\textbf{26.8}$^\dag$\\
      7 & 1 + 4 &10.5$^\dag$&15.5$^\dag$&18.5$^\dag$&22.0$^\dag$&\textbf{11.5}$^\dag$&\textbf{17.9}$^\dag$&\textbf{21.7}$^\dag$&26.4$^\dag$\\
      \hline
      \multicolumn{10}{l}{\textit{Other Baselines for Japanese}}\\
      8 & MultiMASS (Ja) &7.1&12.1&15.1&20.5&6.9&13.0&17.7&24.1\\
      9 & Deshuffling (Ja) &6.8&12.7&16.6&21.0&7.8&14.7&19.3&24.9\\
      10 & 1 + 9 &8.2&13.3&17.0&21.4
      &8.3&15.5&19.5&25.4\\
      \midrule
      \midrule
      \multicolumn{10}{l}{\textit{Proposed methods for English}}\\
      11 & PMASS.P &6.8&12.1&15.9&20.7&5.5&13.5&17.8&24.5\\
      11* & PMASS.S &6.5&12.3&16.2&21.2&6.2&13.5&18.2&24.6\\
      12 & HFSS &10.5$^\dag$&\textbf{16.3}$^\dag$&\textbf{18.9}$^\dag$&\textbf{22.6}$^\dag$&9.8$^\dag$&17.8$^\dag$&\textbf{21.7}$^\dag$&\textbf{26.8}$^\dag$\\
      13 & ENSS &\textbf{11.2}$^\dag$&\textbf{16.7}$^\dag$&\textbf{19.0}$^\dag$&22.1$^\dag$&\textbf{11.7}$^\dag$&\textbf{18.7}$^\dag$&\textbf{22.5}$^\dag$&\textbf{27.0}$^\dag$\\
      %\midrule
      %\multicolumn{10}{l}{\textit{Combinations of proposed methods with MASS}}\\
      %? & 1 + 9 & &&&&&&&&&&&\\
      \hline
      \multicolumn{10}{l}{\textit{Other baselines for English}}\\
      14 & MultiMASS (En) &6.9&12.0&15.2&20.1&7.0&12.8&17.5&23.8\\
      15 & Deshuffling (En) &6.6&12.5&15.9&20.9&6.8&14.1&19.2&24.7\\
      16 & 1 + 15 &7.7&13.2&16.7&21.0&8.6&15.7&20.4&25.6\\
      \midrule
      \midrule
      \multicolumn{10}{l}{\textit{Combination of methods for Japanese and English}}\\
      17 & 4 + 13 &\textbf{10.9}$^\dag$&\textbf{16.4}$^\dag$&18.7$^\dag$&\textbf{22.3}$^\dag$&\textbf{11.9}$^\dag$&\textbf{18.4}$^\dag$&\textbf{22.0}$^\dag$&26.5$^\dag$\\
      18 & 10 + 16 (baseline) &7.2 &12.6&16.4&20.9&8.4&14.8&19.1&25.5\\
      \bottomrule
\end{tabular}
}
\caption{BLEU scores for simulated low/high-resource settings for Japanese--English ASPEC translation using from 3k to 50k parallel sentences for fine-tuning. Pre-trained models used for fine-tuning are numbered according to their description in Section~\ref{sec:pretrainedmodels}. Results better than MASS with statistical significance $p<0.05$ are marked in \dag. Bold denotes the three top scores.} 
\label{ja-en}
 \end{center}
\end{table}

\begin{table}[t]
\begin{center}
\resizebox{0.8\textwidth}{!}{
\begin{tabular}{cc c c c c c c c c }

      \toprule
     \multirow{2}{*}{\#} &\multirow{2}{*}{Model} &\multicolumn{4}{c}{Ja-Zh}&\multicolumn{4}{c}{Zh-Ja}\\
      %\midrule
      &&  3k &  10k & 20k & 50k & 3k & 10k & 20k & 50k \\
      \midrule
      \multicolumn{4}{l}{\textit{Main baselines}}\\
      0 & w/o pre, vanilla &	0.7&3.4&11.5&21.0 &1.9&4.5&16.0&28.2 \\
      0* & w/o pre, optimized & 3.7 & 12.0 & 19.5 & 23.3 & 6.7 & 15.8 & 24.8 & 31.2 \\
      1 &MASS &15.7&20.3&22.4&24.7&19.4&25.9&29.4&32.9\\
      1* & BART (text infilling) & 13.5 & 19.0 & 21.3 & 24.4 & 20.3$^\dag$ & 25.8 & 29.1 & 33.0 \\
      \midrule
      \multicolumn{4}{l}{\textit{Proposed methods}}\\
      2 &BMASS&	16.7$^\dag$&21.1$^\dag$&23.0$^\dag$&25.3$^\dag$ &20.9$^\dag$&27.2$^\dag$&30.2$^\dag$&33.7$^\dag$\\
      3 &BRSS&15.6&21.1$^\dag$&22.6&24.9&20.7$^\dag$&26.8$^\dag$&30.0$^\dag$&33.3$^\dag$\\
      4 &JASS&\textbf{17.1}$^\dag$&\textbf{22.2}$^\dag$&\textbf{23.2}$^\dag$&25.2$^\dag$&21.6$^\dag$&27.5$^\dag$&\textbf{30.4}$^\dag$&\textbf{33.6}$^\dag$\\
      \midrule
      \multicolumn{10}{l}{\textit{Combinations of proposed methods with MASS}}\\
      7&1 + 4&17.0$^\dag$&21.7$^\dag$&23.1$^\dag$&\textbf{25.4}$^\dag$&\textbf{21.8}$^\dag$&\textbf{27.6}$^\dag$&30.2$^\dag$&33.4$^\dag$\\
      \midrule
      \multicolumn{4}{l}{\textit{Other baselines}}\\
      8&MultiMASS&14.5&20.5	&22.3&	24.7&19.6	&25.7&	29.8&	33.2\\
      9&Deshuffling&14.1&	19.5&	21.6&	24.3&18.4	&25.0	&28.7	&32.8\\
      10& 1 + 9 &15.0&20.2&22.1&25.0&18.9&25.9&29.3&33.1\\
      \bottomrule
\end{tabular}
}
\caption{BLEU scores for simulated low-resource settings for Japanese--Chinese ASPEC translation using 3k to 50k parallel sentences for fine-tuning. Results better than MASS with statistical significance $p<0.05$ are marked in \dag.} 
\label{ja-zh}
 \end{center}
\end{table}

\begin{table}[t]
\begin{center}
\resizebox{0.8\textwidth}{!}{
\begin{tabular}{cc c c c c c c c c}

      \toprule
     \multirow{2}{*}{\#} &\multirow{2}{*}{Model} &\multicolumn{4}{c}{Ja-Zh}&\multicolumn{4}{c}{Zh-Ja}\\
      %\midrule
      &&  3k &  10k & 20k & 50k & 3k & 10k & 20k & 50k \\
      \midrule
      \multicolumn{4}{l}{\textit{Main baselines}}\\
      0 & w/o pre, vanilla &0.9&2.9&2.9&6.0&1.6&2.9&3.9&6.5\\
      0* & w/o pre, optimized & 3.3 & 7.8 & 11.7 & 21.9 & 6.7 & 12.0 & 16.2 & 24.2 \\
      1 &MASS & 7.7 &15.4&18.3&23.4&9.6&17.6&23.3&27.1\\
      1* & BART (text infilling) & 5.9  & 14.0 & 18.0 & 21.8 & 8.7  & 17.8 & 24.2$^\dag$ & 28.5$^\dag$ \\
      \midrule
      \multicolumn{4}{l}{\textit{Proposed methods}}\\
      2 &BMASS&10.8$^\dag$&15.7&\textbf{20.1}$^\dag$&24.5$^\dag$&16.2$^\dag$&19.4$^\dag$&25.4$^\dag$&\textbf{30.0}$^\dag$\\
      3 &BRSS&11.6$^\dag$&16.2$^\dag$&20.0$^\dag$ &24.6$^\dag$ &15.7$^\dag$&21.6$^\dag$&25.0$^\dag$&28.3$^\dag$\\
      4 &JASS&\textbf{12.0}$^\dag$&\textbf{17.0}$^\dag$&\textbf{20.1}$^\dag$&\textbf{25.0}$^\dag$&\textbf{16.6}$^\dag$&21.2$^\dag$&\textbf{26.5}$^\dag$&29.2$^\dag$\\
      \midrule
      \multicolumn{10}{l}{\textit{Combinations of proposed methods with MASS}}\\
      7&1 + 4&11.8$^\dag$&16.8$^\dag$&\textbf{20.1}$^\dag$&24.6$^\dag$&\textbf{16.6}$^\dag$&\textbf{22.3}$^\dag$&25.5$^\dag$&29.6$^\dag$\\
      \midrule
      \multicolumn{4}{l}{\textit{Other baselines}}\\
      8&MultiMASS&8.2&13.8&18.6&21.5&10.7&17.3&22.0&26.4\\
      9&Deshuffling&9.3&14.2&18.7&22.7&12.4&18.4&23.2&27.4\\
      10& 1 + 9 &8.7&13.8&19.4&23.2&14.3&18.8&24.8&27.8\\
      \bottomrule
\end{tabular}
}
\caption{BLEU scores for simulated low-resource settings for Japanese--Chinese Wikipedia translation using from 3k to 50k parallel sentences for fine-tuning. Results better than MASS with statistical significance $p<0.05$ are marked in \dag.} 
\label{ja-zh2}
 \end{center}
\end{table}

\begin{table}[t]
\begin{center}
%\arrayrulecolor{blue}
%\resizebox{0.8\textwidth}{!}{
\begin{tabular}{cc c c c c}

      \toprule
      \multirow{2}{*}{\#} &\multirow{2}{*}{Model} &\multicolumn{2}{c}{En-Ko}&\multicolumn{2}{c}{Ko-En}\\
      %\midrule
      & & 20k & 94k & 20k & 94k \\
      \midrule
      \multicolumn{4}{l}{\textit{Main baselines}}\\
      0 & w/o pre-training & 1.3 & 2.1 & 2.9 & 4.5\\
      0* & optimized transformer & 2.1 & 3.7 & 3.9 & 8.3 \\
      1 &MASS & 2.9 & 4.5 & 5.6 & 9.6 \\
      \midrule
      \multicolumn{4}{l}{\textit{Proposed methods}}\\
      2 & PMASS & 2.4 & 4.3 & 5.2 & 9.4 \\
      3 & HFSS & 3.1 & 4.8 & \textbf{7.7}$^\dag$ & 10.3$^\dag$ \\
      4 & ENSS (1 + 3) & \textbf{3.2} & \textbf{5.0}$^\dag$ & 6.8$^\dag$ & \textbf{10.9}$^\dag$ \\
      \midrule
      \multicolumn{4}{l}{\textit{Other combinations}}\\
       & 2 + 3 & 3.0 & 4.7 & 7.0$^\dag$ & 10.6$^\dag$ \\
      \bottomrule
\end{tabular}
%}
\caption{BLEU scores for simulated low-resource settings for English--Korean News translation using 20k and 94k parallel sentences for fine-tuning. Results better than MASS with statistical significance $p<0.05$ are marked in \dag. The BLEU scores are relatively low because English--Korean is a dissimilar language pair. \citet{sennrich-zhang-2019-revisiting} and \citet{park-etal-2016-korean} reported similar BLEU results.}
\label{en-ko}
 \end{center}
\end{table}

In Tables~\ref{ja-en} and~\ref{ja-zh}, where we simulate several low-resource settings for Japanese--English and Japanese--Chinese translations on ASPEC with different pre-training datasets; in Table~\ref{ja-zh2} and~\ref{en-ko}, where we use realistic low-resource settings for Wikipedia Japanese--Chinese translation and News English--Korean translation, we observe that all settings using pre-training outperform those without pre-training (\#0 \& \#0*), which indicates the importance of pre-training. The results also indicate that JASS (\#4) and ENSS (\#13) are generally better than MASS (\#1). With regard to two main baselines with pre-training, MASS and BART (text infilling), we observe that MASS outperforms BART (text infilling) in most cases as shown in Table~\ref{ja-en},~\ref{ja-zh},~\ref{ja-zh2}. So we focus on the comparisons with MASS in the following analyses. We also present the results by combing BART (text infilling) with ours in Appendix~\ref{appb}.\footnote{Note that better results from BART than MASS in~\citet{lewis-etal-2020-bart} are based on the multi-task objectives while we are comparing with the most effective single task within BART here.} Without pre-training, we observe that using optimized transformer (\#0*) benefits the low-resource setting, which has been proven by \citet{araabi-monz-2020-optimizing, sennrich-zhang-2019-revisiting}. However, pre-training can further improve the optimized baselines without pre-training.

Particularly, for the Japanese--English translation, BMASS (\#2) is comparable to MASS; BRSS (\#3 \& \#3(R)) and their combination, along with JASS (\#5) are significantly better than MASS. However, as summarized in Tables~\ref{ja-zh} and~\ref{ja-zh2}, the results for two parallel corpora on different domains for Japanese--Chinese yield significantly better results when using our proposed BMASS and BRSS. We observe that only a few settings on Japanese-to-Chinese BRSS yield lower BLEU results than MASS, whereas other settings using the proposed methods yield better results than MASS by significant margins. Although MASS is better than BMASS for Japanese--English translation, the reverse can be observed for the Japanese--Chinese translation. This indicates that the effects of the proposed span-masking techniques might correlate with specific translation directions and domains. We suppose it is worth exploring the span-masking tricks that are non-sensitive to language pairs and domains in the future.

As summarized in Table~\ref{ja-en} and~\ref{en-ko}, our proposed methods of leveraging linguistic knowledge for English yield significantly higher BLEU results when we perform the reordering pre-training task, HFSS (\#12). However, the proposed linguistically-driven masked language model PMASS.P (\#11) and PMASS.S (\#11*) yielded comparable results to several other baseline methods such as MultiMASS (\#14) and deshuffling (\#15). This demonstrates that the syntactical span-based masked language model may merely work on head-final languages such as Japanese.\footnote{The weak performance of PMASS can also be attributed to the discrete nature of the remaining tokens (tokens that are not masked) without constituting complete semantic spans. We will attempt chunking-based masking for PMASS in future work to allow PMASS to be performed in a manner similar to BMASS.} Considering the weak performance of the PMASS, we combined HFSS with MASS for ENSS. The multi-task pre-trained ENSS yielded the highest results on almost all the low-resource settings. We will explore proper chunking techniques for linguistically-driven span-masking pre-training for languages like English in the future.

However, in Table~\ref{ja-en}, when performing a universal linguistically-driven pre-training simultaneously for Japanese and English (\#17), we did not achieve further significant BLEU improvements. This can be attributed to the increased dependence of NMT on specific linguistic information on a single language side, and the joint pre-training does not allow linguistic knowledge transfer across languages and between dissimilar languages.

In addition to the main baseline MASS, we also performed several other sequence-to-sequence pre-training baselines: MultiMASS (\#8 \& \#14) and deshuffling (\#9 \& \#15) along with their multi-task combinations (\#10, \#16 \& \#18) for Japanese and English. As summarized in Tables~\ref{ja-en},~\ref{ja-zh},~\ref{ja-zh2}, and~\ref{en-ko}, we observe that the proposed masked style pre-training task, BMASS, and reordering pre-training tasks, BRSS \& HFSS, outperform these baselines by significant margins, thereby indicating that linguistically-driven methods should be superior to self-supervised pre-training without leveraging linguistic features. Moreover, we investigated the percentages of the words of which the position changed. For Japanese pre-training, the percentages are 79.58\% for BRSS and 94.72\% for deshuffling. For English pre-training, the percentages are 91.97\% for HFSS and 95.22\% for deshuffling. Although there exists a gap for the percentages between BRSS and deshuffling, we can see that the percentages of deshuffling and HFSS are similar, which demonstrates that the quality of the linguistically generated reordered sentence is much more important than the percentage.

As summarized in Table~\ref{ja-en}, BRSS-F (English-order to Japanese-order) yielded slightly better results than BRSS-R (vice-versa); thus, we only experimented with BRSS-F for the remaining experiments. We suppose that the reason is that training the decoder with the original sentence is more important than training the encoder with it, which is also the reason why BART pre-training~\cite{lewis-etal-2020-bart} treats the original sentence as the target sentence to be predicted from the decoder.\footnote{In BART, the original sentences without any noise are treated as target sentences.} In other words, forcing the decoder to generate a natural sentence leads to a better initialized decoder for NMT. Meanwhile, HFSS pre-training is performed in an analogous manner for the same reason.

As mentioned above, JASS yields the best results when we consider only linguistically driven methods for Japanese. After combining the proposed methods for Japanese with MASS (\#5$\sim$\#7 in Table~\ref{ja-en}), we observe comparable results as compared to JASS by combing MASS and BRSS. This indicates the effects of combining masked style methods and reordering style methods. In Table~\ref{ja-zh} and~\ref{ja-zh2}, we believe that BMASS is better than MASS for combining with BRSS because of the significant improvements yielded by BMASS.
%no significant improvements were observed. This demonstrates that linguistic-aware methods can substitute linguistic-agnostic methods.

Moreover, as summarized in Tables~\ref{ja-zh} and~\ref{ja-zh2}, we observe that on the ASPEC domain, JASS improves up to 2.2 BLEU scores, whereas on the Wikipedia domain, JASS achieves up to 7.0 BLEU improvements. This demonstrates the promising performance of the proposed methods. Meanwhile, this indicates that the overlapping of pre-training domain with the fine-tuning domain is directly proportional to the realization of improvements by linguistically-driven pre-training methods. 

Finally, by comparing the BLEU results in Table~\ref{ja-en} with those reported by~\citet{mao-etal-2020-jass}, we find that the BLEU scores of models pre-trained with News Crawl are better than those pre-trained with the Common Crawl monolingual corpus, which shows that pre-training with a high-quality monolingual dataset leads to superior fine-tuning results.

\subsection{Adequacy Evaluation}
\begin{table}[t]
\begin{center}
\arrayrulecolor{black}
%\resizebox{\columnwidth}{!}{
\begin{tabular}{cc |cc|c  c | c c | cc}

      \toprule
   \multirow{2}{*}{\#}&
       \multirow{2}{*}{Model}& \multicolumn{2}{c|}{ASPEC}& \multicolumn{2}{c|}{ASPEC}&  \multicolumn{2}{c|}{Wikipedia} & \multicolumn{2}{c}{News}\\
       &&Ja-En&En-Ja&Ja-Zh&Zh-Ja&Ja-Zh&Zh-Ja&En-Ko&Ko-En\\
      \midrule
      *& Reference &\multicolumn{2}{c|}{80.78} &\multicolumn{2}{c|}{86.10}&\multicolumn{2}{c|}{87.26}&\multicolumn{2}{c}{73.93}\\
      \midrule
      0& w/o pre-training &52.59&45.89&69.54&67.08&57.55&56.48 & 59.68 & 65.38 \\
      \midrule
      1& MASS &75.63&76.09&85.52&86.32&81.08&78.52 & 72.54 & 73.30 \\
      \midrule
      2&BMASS&75.75&76.68&85.42&86.49&80.91&81.36&-&-\\
      3&BRSS&78.34&76.66&85.87&86.54&81.71&\textbf{84.29}&-&-\\
      4&JASS &\textbf{80.00}&77.63&\textbf{85.96}&\textbf{86.58}&\textbf{85.39}&83.08&-&-\\
      \midrule
      11&PMASS&76.08&73.67&-&-&-&-&71.90&74.14\\
      12&HFSS&79.38&79.13&-&-&-&-&73.60&75.59\\
      13&ENSS&79.79&\textbf{79.64}&-&-&-&-&\textbf{74.13}&\textbf{75.66}\\
      \bottomrule
\end{tabular}
%}
\caption{Adequacy scores evaluated by LASER embedding-based cosine similarity for ASPEC Japanese--English, Japanese--Chinese, Wikipedia Japanese--Chinese and News English--Korean translations, respectively, using 10k sentences for fine-tuning (using 94k sentences for English--Korean). Reference (*) is the cosine similarity between test sets in two languages.} 
\label{laser}
 \end{center}
\end{table}
Reference-free MT evaluation evaluates the translation system without using the target reference. Such an evaluation can help circumvent the noise existing in the references of translation targets. After the emergence of multilingual sentence encoders~\cite{artetxe-schwenk-2019-massively}, \citet{yankovskaya-etal-2019-quality} proposed the use of multilingual sentence embeddings encoded by LASER to implement the reference-free MT evaluation. More precisely, we first apply LASER to encode the source sentence and the translated sentence, respectively. Thereafter, the cosine value of those two embeddings is used to evaluate the similarity between the source and translation. This cosine value is thus the metric used to evaluate translation adequacy. This approach has two advantages. The first advantage is that target references are not required, as mentioned above. The other advantage is that every two translation directions can be compared with each other because language-agnostic embedding is used for evaluation.

We report the adequacies in Table~\ref{laser}. First, we observe that methods with pre-training can yield more semantically correct translations than those without pre-training. Second, our proposed methods can significantly obtain higher LASER similarity scores than the MASS baseline, particularly the results on ASPEC Japanese--English, Wikipedia Chinese--Japanese and News English--Korean translations. Moreover, we can observe that the adequacy results obtained from the LASER embedding-based cosine similarity scores are consistent with the BLEU results.

\begin{table}[t]
\begin{center}
%\resizebox{\columnwidth}{!}{
\begin{tabular}{cc |cc|c  c | c c}

      \toprule
   \multirow{2}{*}{\#}&
       \multirow{2}{*}{Model}& \multicolumn{2}{c|}{BLEU}& \multicolumn{2}{c|}{Adequacy}&  \multicolumn{2}{c}{Fluency}\\
       &&Ja-Zh&Zh-Ja&Ja-Zh&Zh-Ja&Ja-Zh&Zh-Ja\\
      \midrule
      0& w/o pre-training &2.9&2.9&1.22&1.05&3.90&3.99\\
      \midrule
      1& MASS &15.4&17.6&2.72&2.33&4.11&4.09\\
      \midrule
      2&BMASS &15.7&19.4&3.12&2.88&4.34&4.32\\
      3&BRSS &16.2&\textbf{21.6}&3.30&3.35&4.30&\textbf{4.40}\\
      4&JASS &\textbf{17.0}&21.2&\textbf{3.79}&\textbf{3.44}&\textbf{4.47}&4.36\\
      \bottomrule
\end{tabular}
%}
\caption{Adequacy and fluency of Wikipedia Japanese--Chinese translations using 10k sentences for fine-tuning.} 
\label{human}
 \end{center}
\end{table}

\begin{table}[t]
\begin{center}
%\resizebox{\columnwidth}{!}{
\begin{tabular}{cc |cc|c  c | c c}

      \toprule
   \multirow{2}{*}{\#}&
       \multirow{2}{*}{Model}& \multicolumn{2}{c|}{BLEU}& \multicolumn{2}{c|}{Adequacy}&  \multicolumn{2}{c}{Fluency}\\
       &&Ja-En&En-Ja&Ja-En&En-Ja&Ja-En&En-Ja\\
      \midrule
      0& w/o pre-training &2.1&2.7&1.08&1.08&2.56&3.60\\
      \midrule
      1& MASS &13.8&16.0&2.61&3.03&3.40&4.17\\
      \midrule
      2&PMASS &12.1&13.5&2.40&2.76&3.24&4.07\\
      3&HFSS &16.3&17.8&3.24&4.00&3.60&4.31\\
      4&ENSS &\textbf{16.7}&\textbf{18.7}&\textbf{3.72}&\textbf{4.11}&\textbf{3.76}&\textbf{4.42}\\
      \bottomrule
\end{tabular}
%}
\caption{Adequacy and fluency of ASPEC Japanese--English translations using 10k sentences for fine-tuning.} 
\label{human2}
 \end{center}
\end{table}

\subsection{Human Evaluation}

Following~\citet{nakazawa-etal-2018-overview}, we performed adequacy and fluency evaluations for the Japanese--Chinese and Japanese--English translations when 10k Wikipedia parallel sentences and 10k ASPEC parallel sentences were used for fine-tuning the pre-trained models. We randomly sampled 100 test-set English sentences and blindly evaluated their translations across various models. Each sentence was scored on a scale of 1 to 5, with 1 representing the worst score. The higher the score, the more adequate (meaningful) or fluent (well-formed) the sentence is. The final score was the average of the scores of 100 sentences. We did not consider the references, but only considered the sources for our evaluation.

In Table~\ref{human} and~\ref{human2}, we can observe that NMT models, even without pre-training, are capable of generating rather fluent sentences, and the lack of parallel sentences (low-resource scenario) will mainly influence the translation adequacy (refer to the extremely low adequacy of models without pre-training). Meanwhile, we can observe that our proposed BMASS, BRSS, JASS, HFSS, and ENSS result in large improvements in adequacy and moderate improvements in fluency, for both translation directions, whereas PMASS yielded marginal improvements. The improved performance of adequacy compared with that of MASS demonstrates the effectiveness of linguistically-driven pre-training methods. Moreover, we can observe that the results of human evaluation are almost consistent with those of BLEU.

\begin{table}[t]
    \centering
    \resizebox{\columnwidth}{!}{ 
    \begin{tabular}{lcl}
    \toprule
    \vspace{1.5mm}
    \multirow{2}{*}{\#}&Reference--Ja & \begin{tabular}{l}
         \begin{CJK}{UTF8}{ipxm}水の性質の多様性について,まず,水分子同士の間に働く力である水素結合と,そのネットワーク構造\end{CJK}\\\begin{CJK}{UTF8}{ipxm}について解説した。\end{CJK} \end{tabular}\\
    &Reference--En & \begin{tabular}{l} Various properties of water were explained on hydrogen bonds in which the force works among the water \\ molecules and the network structure. \end{tabular}\\
    \midrule\midrule
    %\vspace{1mm}
    %\multicolumn{3}{l}{\textit{Japanese to English translation fine-tuned by 10k ASPEC-JE parallel sentences}} \\
    \vspace{1.5mm}
    0&w/o pre-training & \begin{tabular}{l}This paper introduces the outline of the development of the system, and it is described. \end{tabular}\\
    \vspace{1.5mm}
    1&MASS & \begin{tabular}{l}The network structure of the water, hydrogen combination as the power of the water, and the network str-\\ucture are explained. \end{tabular}\\
    \vspace{1.5mm}
    2&BMASS & \begin{tabular}{l}On the basis of the water properties, hydrogen coupling and the network structure are explained in the fir-\\st stage of water. \end{tabular}\\
    \vspace{1.5mm}
    3&BRSS & \begin{tabular}{l}On the formation of the water, this paper explains hydrogen bond and hydrogen bond, which is connected \\ between the water vapor man fellows. \end{tabular}\\
    \vspace{1.5mm}
    4&JASS & \begin{tabular}{l}This paper explains the development of the properties of water and hydrogen combination, which is the \\ power between the moisture man fellows and the network structure. \end{tabular}\\
    \vspace{1.5mm}
    8&MultiMASS (Ja) & \begin{tabular}{l}This paper explains the rich characteristics of the water and also explains the network structure of the hyd-\\rogen joining with the network structure. \end{tabular}\\
    \vspace{1.5mm}
    9&Deshuffling (Ja) & \begin{tabular}{l}This paper explains the potential of the water in the water, and the network structure that is connected be-\\tween the water and hydrogen joining. \end{tabular}\\
    \vspace{1.5mm}
    10 & Multi-task baseline (Ja) &
    \begin{tabular}{l}The active properties of water are explained, and hydrogen combination that is connected to the network \\ structure and the power of the water are explained. \end{tabular}\\
    \vspace{1.5mm}
    11&PMASS & \begin{tabular}{l}The importance of the properties of water and the network structure, which is the active component of the \\ water, are explained. \end{tabular}\\
    \vspace{1.5mm}
    12&HFSS & \begin{tabular}{l}The formation of the properties of water is first explained, then hydrogen combination and the network str-\\ucture between the moisture man. \end{tabular}\\
    \vspace{1.5mm}
    13&ENSS & \begin{tabular}{l}The importance of the property of the water is first explained: hydrogen combination and the network str-\\ucture, which is the power for the entire body of the water.\end{tabular}\\
    \vspace{1.5mm}
    14&MultiMASS (En) & \begin{tabular}{l}The growth of the water is explained, and the network structure and structure are explained through the \\ hydrogen combination and network structure.\end{tabular}\\
    \vspace{1.5mm}
    15&Deshuffling (En) &
    \begin{tabular}{l}The network structure of the water properties is explained, and the network structure with hydrogen in \\ the water is described. \end{tabular}\\
    16 & Multi-task baseline (En) &
    \begin{tabular}{l}This paper explains the growth of the water properties, and it also explains hydrogen bonding and its net-\\work structure with the ability to develop between the water molecules.\end{tabular}\\
    \bottomrule
    \end{tabular}
    }
    \caption{Japanese--English translation examples fine-tuned using 10k ASPEC parallel sentences.}
    \label{case1}
\end{table}

\begin{table}[t]
    \centering
    \resizebox{\columnwidth}{!}{ 
    \begin{tabular}{lcl}
    \toprule
    \vspace{1.5mm}
    \multirow{2}{*}{\#}&Reference--Ja & \begin{tabular}{l}
         \begin{CJK}{UTF8}{ipxm}翌嘉永２年（１８４９年）１１月２２日に婚姻し、御廉中様（将軍世子の正室）と称された。\end{CJK} \end{tabular}\\
    &Reference--Zh & \begin{CJK}{UTF8}{gbsn}翌嘉永２年（１８４９年）１１月２２日成婚，被称作御廉中样（将军世子的正室）。\end{CJK} \\
    & & (He married on November 22nd in the second year of Kaei (1849), and was called Gorenju (Seishitsu of the Shogun's son).)\\
    \midrule\midrule
    %\multicolumn{3}{l}{\textit{Chinese to Japanese translation fine-tuned using 10k Wikipedia parallel sentences}} \\
    0&w/o pre-training & \begin{CJK}{UTF8}{ipxm}慶長２年（１５８３年）、劉宮将軍となり、幕府の命を率いた。\end{CJK} \\
    \vspace{1.5mm}
    &&(He became \begin{CJK}{UTF8}{ipxm}劉宮 \end{CJK} Shogun in the second year of Keicho, and lived the life of Shogun.)\\
    1&MASS & \begin{CJK}{UTF8}{ipxm}翌嘉永２年（１８４９年）５月２７日に成婚。\end{CJK} \\
    \vspace{1.5mm}
    &&(He married on May 27th in the second year of Kaei (1849).)\\
    2&BMASS & \begin{CJK}{UTF8}{ipxm}翌嘉永２年（１８４９年）１１月２３日に結婚。\end{CJK} \\
    \vspace{1.5mm}
    &&(He married on November 23rd in the second year of Kaei (1849).)\\
    3&BRSS & \begin{CJK}{UTF8}{ipxm}翌嘉永２年（１８４９年）１１月２３日に成婚し、御廉の中样（将軍世子の正室）と呼ばれる。\end{CJK} \\
    \vspace{1.5mm}
    &&(He married on November 23rd in the second year of Kaei (1849), and was called Naka of Goren (Seishitsu of the Shogun's son).)\\
    4&JASS & \begin{CJK}{UTF8}{ipxm}翌嘉永２年（１８４９年）１１月２２日成婚し、御廉中様（将軍世子の正室）と称した。\end{CJK} \\
    \vspace{1.5mm}
    &&(He married on November 22nd in the second year of Kaei (1849), and was called Gorenju (sSeishitsu of the Shogun's son).)\\
    8&MultiMASS & \begin{CJK}{UTF8}{ipxm}翌嘉永２年（１８４９年）１月２９日に成婚した。\end{CJK} \\
    \vspace{1.5mm}
    &&(He married on January 29th in the second year of Kaei (1849).)\\
    9&Deshuffling & \begin{CJK}{UTF8}{ipxm}翌嘉永２年（１８４９年）１１月２３日、御廉中の正室（将軍世子の正室）と称された。\end{CJK} \\
    \vspace{1.5mm}
    &&(He was called Seishitsu of Gorenju (Seishitsu of the Shogun's son) on November 22nd in the second year of Kaei (1849).)\\
    10 & Multi-task baseline &
    \begin{CJK}{UTF8}{ipxm}翌嘉永２年（１８４９年）１１月２２日成婚し、御廉の代わりに義子と呼ばれる。\end{CJK} \\
    &&(He married on November 22nd in the second year of Kaei (1849), and was called Gishi instead of Goren.)\\
    \bottomrule
    \end{tabular}
    }
    \caption{Japanese--Chinese translation examples fine-tuned using 10k Wikipedia parallel sentences. Sentences in brackets correspond to English sentences of the above Japanese translations.}
    \label{case2}
\end{table}

\begin{table}[t]
\begin{center}
\begin{tabular}{cc cccc }

      \toprule
      \#&Model&  Overall& MASS &BMASS&BRSS\\
      \midrule
      1 &MASS &69.75			&	69.75	&	-	&-\\
      \midrule
      2 &BMASS&77.32	&	-	&	77.32&	-\\
      3 &BRSS&87.90		&	-	&	-	&95.90\\
      4& JASS&85.15	&	-&		77.34	&\textbf{97.89}\\
      \midrule
      5&1 + 2 &	74.53	&	\textbf{70.17}	&	78.59	&-\\
      6&1 + 3&	81.58	&	69.72&		-&	94.43\\
     7& 1 + 4& 80.81	&	\textbf{70.22}		&77.90	&\textbf{97.73}\\
      \bottomrule
\end{tabular}
\caption{Component-wise and overall pre-training accuracies on ASPEC Japanese development sentences. Column names ``MASS,'' ``BMASS,'' and ``BRSS'' denote the pre-training components in the respective model. Note the boost of BRSS accuracy in multitask settings, although the opposite could have been expected.} 
\label{acc1}
\end{center}
\end{table}

\begin{table*}[t]
\begin{center}
\begin{tabular}{cc cccc }

      \toprule
      \#&Model&  Overall&MASS&PMASS&HFSS\\
      \midrule
      1 &MASS &		70.97	&	70.97	&	-	&-\\
      \midrule
      2 &PMASS& 71.04	&	-	& 71.04	&	-\\
      3 &HFSS&	96.48	&	-	&	-	& 96.48 \\
      4& ENSS& 84.97	& \textbf{71.24}	&		-	&\textbf{98.05}\\
      \bottomrule
\end{tabular}
\caption{Component-wise and overall pre-training accuracies on ASPEC English development sentences. Column names ``MASS,'' ``PMASS,'' and ``HFSS'' denote the pre-training components in the respective model. Note the boost of the HFSS accuracy in multitask settings, although the opposite could have been expected.} 
\label{acc2}
\end{center}
\end{table*}

\subsection{Case Study}

We conducted case studies on Japanese-to-English translation fine-tuned using 10k ASPEC parallel sentences and Chinese-to-Japanese translation fine-tuned using 10k Wikipedia parallel sentences to make improvements shown by BLEU score evaluations visible. As summarized in Tables~\ref{case1} and~\ref{case2}, we find that the vanilla NMT system trained using 10k parallel sentences without pre-training can hardly implement the translation. With regard to models with pre-training, we observed that MASS and other baseline models generated several incorrect tokens in terms of semantics, whereas the entire sentence seemed fluent. However, our proposed methods can generate sentences with superior adequacy and fluency, where fewer missing keywords are observed.

\subsection{Pre-training Accuracy}

Pre-training accuracy is the accuracy of the monolingual pre-training tasks, and it can be an indicator of task complexity and pre-training objective performance.
Tables~\ref{acc1} and~\ref{acc2} summarize the component-wise and overall pre-training accuracies for various models, respectively, on the ASPEC Japanese and English development set sentences. Regarding individual component methods, it can be observed that MASS and PMASS are the harder tasks, given their low accuracy, whereas BRSS and HFSS are the easier tasks. Moreover, for Japanese, the accuracy of MASS and BRSS improves when coupled with BMASS, whereas for English, the accuracy of HFSS and MASS improves when they are combined with each other. Cross-referencing these accuracies with the BLEU scores in Table~\ref{ja-en}, we observe that an increase in BLEU scores has no significant relationship with the pre-training accuracy. However, masked language model-based pre-training methods (MASS \& BMASS) seem to act as an accuracy improving catalyst for BRSS and HFSS, and this in turn has a positive impact on the translation quality.

One possible reason for this is that multi-task training of different pre-training methods helps boost the performance of individual methods. This is in accordance with several previous studies on multi-task training for NMT~\cite{dong-etal-2015-multi,lewis-etal-2020-bart,DBLP:journals/tacl/LiuGGLEGLZ20,DBLP:journals/jmlr/RaffelSRLNMZLL20}. Therefore, we recommend that such an analysis of multi-objective pre-training methods can help isolate the importance of individual pre-training objectives. Nevertheless, our analyses reveal that the components of JASS, BMASS, and BRSS, and the components of ENSS, MASS, and HFSS are certainly responsible for improving translation quality for Japanese-involved or English-language pairs.

\subsection{Results in Middle/High-resource Scenarios}
\label{appendix:high-resource}

\begin{table}[t]
\begin{center}
\resizebox{1\textwidth}{!}{
\begin{tabular}{cc c c c c c ccccc}

      \toprule
     \multirow{3}{*}{\#} &\multirow{3}{*}{Model} &\multicolumn{2}{c}{Ja-En}&\multicolumn{2}{c}{En-Ja}&\multicolumn{3}{c}{Ja-Zh}&\multicolumn{3}{c}{Zh-Ja}\\
      %\midrule
      &&\begin{tabular}{c}ASP \\ 200k\end{tabular}&\begin{tabular}{c}ASP \\ 1M\end{tabular}&\begin{tabular}{c}ASP \\ 200k \end{tabular} &\begin{tabular}{c}ASP \\ 1M\end{tabular}&\begin{tabular}{c}ASP \\ 200k\end{tabular} & \begin{tabular}{c}ASP \\ 672k\end{tabular}&\begin{tabular}{c}Wiki \\ 258k\end{tabular}  &\begin{tabular}{c} ASP \\ 200k\end{tabular} &\begin{tabular}{c}ASP \\ 672k\end{tabular}&\begin{tabular}{c} Wiki \\ 258k\end{tabular} \\
      \midrule
      \multicolumn{4}{l}{\textit{Main baselines}}\\
      0 &w/o pre-training &26.1&27.5&33.4&35.8&27.0&31.2&24.6&36.7&42.4&30.4\\
      1 &MASS &26.5&28.8&33.7&37.6&27.2&\textbf{33.0}&29.4&36.7&\textbf{44.8}&34.6\\
      \midrule
      \multicolumn{4}{l}{\textit{Proposed methods for Japanese}}\\
      2 &BMASS&26.5&\textbf{28.9}&33.8&37.8&\textbf{27.8}&32.6&29.8&36.4&44.7&\textbf{35.6}\\
      3 &BRSS&\textbf{26.8}&28.4&\textbf{34.0}&37.4&27.2&32.7&30.8&36.8&44.5&35.0\\
      4 &JASS&26.7&28.8&33.2&37.5&27.2&32.7&\textbf{31.1}&\textbf{37.4}&\textbf{44.8}&35.4\\
      \midrule
      \multicolumn{4}{l}{\textit{Proposed methods for English}}\\
      11&PMASS.P&26.3&28.0&33.5&37.0&-&-&-&-&-&-\\
      11*&PMASS.S&26.2&28.9&33.2&37.8&-&-&-&-&-&-\\
      12& HFSS &26.5&28.6&33.9&37.7&-&-&-&-&-&-\\
      13& ENSS&26.3&28.8&33.9&\textbf{37.9}&-&-&-&-&-&-\\
      \bottomrule
\end{tabular}
}
\caption{BLEU scores in middle/high-resource scenarios. ``ASP'' and ``Wiki'' denote ASPEC and Wikipedia parallel corpus, respectively.} 
\label{high-resource}
 \end{center}
\end{table}

As summarized in Table~\ref{high-resource}, we report that BLEU leads to middle/high-resource scenarios. The fine-tuning is performed by more than 200k parallel sentences on the respective language pair and domain. By comparing with models without pre-training, we find that pre-training can still lead to some improvements, but much less than those in low-resource scenarios. Second, we observe that most pre-training methods obtained comparable BLEU results regardless of whether they were linguistically-driven methods or not. This indicates that in middle/high-resource scenarios, our proposed methods might be limited, which also shows that linguistically-driven supervision can be utilized to compensate for the lack of parallel sentences.

\section{Conclusion}
In this study, we proposed JASS and ENSS pre-training methods that leverage information from syntactic structures of sentences on the basis of language-agnostic pre-training schemes such as MASS for NMT. Our work leveraged abundant monolingual data and syntactic analysis such that the pre-training phase became aware of specific language structures. Our experiments on ASPEC Japanese--English, Japanese--Chinese, Wikipedia Japanese--Chinese, and News English--Korean translations demonstrated that JASS and ENSS outperform MASS and other language-agnostic pre-training methods in most low-resource settings. %Furthermore, we demonstrated that JASS and ENSS can completely substitute the corresponding language-agnostic pre-training tasks and enhance the performance of low-resource NMT.
This demonstrates the importance of injecting language-specific information into the pre-training objective, as well as the benefit of multi-task pre-training with masked style and reordering objectives. Our adequacy evaluation through LASER, human evaluation, and case study also demonstrated that our methods resulted in a significant improvement in terms of the adequacy and fluency of translations. The analyses of pre-training accuracy reveal the complementary nature of individual tasks within JASS and ENSS.

Our future work will focus on implementing linguistic-aware multilingual pre-training using more languages for more robust pre-trained models. We also note that~\citet{DBLP:journals/jmlr/RaffelSRLNMZLL20} demonstrated that several NLP tasks such as text understanding can be reformulated as text-to-text tasks. This broadens the domain of usefulness of sequence-to-sequence pre-training tasks including ours, and we will be interested in evaluating our approach on various NLP tasks.

\begin{acks}
We sincerely thank Dr. Raj Dabre, Dr. Fabien Cromieres, and Mr. Haiyue Song for their support and insightful comments on the JASS part of this work. This work was supported by Grant-in-Aid for Young Scientists \#19K20343, JSPS and Information/AI/Data Science Doctoral Fellowship of Kyoto University.
\end{acks}

\bibliographystyle{ACM-Reference-Format}
\bibliography{reference}

\appendix
\section{Algorithms for PMASS}
\label{algo}
In this section, we introduce Algorithms~\ref{PMASS.S} and~\ref{PMASS.P} for PMASS.S and PMASS.P respectively. We utilize the HPSG parsing result (Figure~\ref{hf} (left)) to detect phrase spans to be masked. For PMASS.S, we can rapidly detect an entire phrase span to be masked. For PMASS.P, we start from the root of the HPSG parsing tree and stochastically mask the left child or the right child; then shift to the unmasked child node to find the next masking candidate. We implement this in a recursive manner.

\section{Hyperparameters for Optimized Transformer}
\label{appa}

Following~\citet{araabi-monz-2020-optimizing}, we use the hyperparameter settings shown in Table~\ref{hyperp} for training optimized transformer on different parallel data settings. Although optimized hyperparameter settings can significantly improve low-resource NMT, they require laborious grid search for the optimal setting while fine-tuning NMT based on pre-trained models do not.

\section{Results of Combining BART with Ours}
\label{appb}
In Table~\ref{ja-en-bart},~\ref{ja-zh-bart} and~\ref{ja-zh2-bart}, we report the results of combining BART and our proposed methods for Japanese--English and Japanese--Chinese translations. We observe that BART (text infilling) can not further improve our proposed methods, which indicates that BART (text infilling) does not have a complement nature with our linguistically-driven multi-task pre-training methods.

\begin{algorithm}[h]
\SetAlgoLined
\KwIn{Length of the sentence \textit{L}, tree of HPSG parsing result for the sentence \textit{T}.}
\KwOut{Token List \textit{M} consisting of all the tokens on \textit{N}. (to be masked)}
Initialize Current Node \textit{N} by \textit{ROOT} of \textit{T}\;
\While{number of tokens on $N > int(L / 2)$}{
\eIf{number of tokens on left child of $N >$ number of tokens on right child of $N$}{$N$ $\gets$ \textit{left child of} $N$\;}{$N$ $\gets$ \textit{right child of} $N$\;}
}
\caption{Algorithm for determining masked phrase spans for PMASS.S.}
\label{PMASS.S}
\end{algorithm}

\begin{algorithm}[htpb]
\SetAlgoLined
\SetKwFunction{FMain}{Pmass}
\SetKwProg{Fn}{Function}{:}{}
\KwIn{Length of the sentence \textit{L}, tree of HPSG parsing result for the sentence \textit{T}.}
\KwOut{\FMain{N=\textit{ROOT} of \textit{T}, L, l=0, M=empty list} (tokens to be masked)}
\Fn{\FMain{$N$, $L$, $l$, $M$}}{
\uIf{tag of N is sentence}{\textbf{return} \FMain{child of N, L, l, M}}
\uElseIf{tag of N is tok}{
\uIf{$int(L/2) - l > 0$}{\textit{M.append(token on N)}}
\textbf{return} \textit{M}
}
\uElseIf{N only has one child and N.tag is cons}{\textbf{return} \FMain{child of N, L, l, M}}
\uElse{$ll$ $\gets$ \textit{number of tokens on the left child of }$N$\;
$lr$ $\gets$ \textit{number of tokens on the right child of }$N$\;
\uIf{ll is 1 and lr is 1}{
\uIf{$int(L/2) - l > 1$}{\textit{M.append(token on N)}}
\textbf{return} \textit{M}
}
\uElseIf{$int(L/2)<=l$}{\textbf{return} \textit{M}}
\uElseIf{$ll<=int(L/2)-l$ and $lr>int(L/2)-l$}{
\uIf{$random$ $p < 0.5$}{$M$ $\gets$ $M +$ \textit{tokens on the left child of N}\; $l$ $\gets$ $l+ll$\; \textbf{return} \FMain{right child of N, L, l, M}}
\uElse{\textbf{return} \FMain{right child of N, L, l, M}}
\textbf{end}
}

\uElseIf{$lr<=int(L/2)-l$ and $ll>int(L/2)-l$}{
\uIf{$random$ $p < 0.5$}{$M$ $\gets$ $M +$ \textit{tokens on the right child of N}\; $l$ $\gets$ $l+lr$\; \textbf{return} \FMain{left child of N, L, l, M}}
\uElse{\textbf{return} \FMain{left child of N, L, l, M}}
\textbf{end}
}
\uElseIf{$ll>int(L/2)-l$ and $lr>int(L/2)-l$}{
\uIf{$random$ $p < 0.5$}{\textbf{return} \FMain{left child of N, L, l, M}}
\uElse{\textbf{return} \FMain{right child of N, L, l, M}}
\textbf{end}
}
\uElse{$M$ $\gets$ $M +$ \textit{tokens on the left child of N}\; $l$ $\gets$ $l+ll$\; \textbf{return} \FMain{right child of N, L, l, M}}

}
}
\textbf{End Function}
\caption{Algorithm for determining masked phrase spans for PMASS.P.}
\label{PMASS.P}
\end{algorithm}

\begin{table}[h]
\begin{center}
%\arrayrulecolor{blue}
%\resizebox{0.9\textwidth}{!}{
\begin{tabular}{crrrrrr}

      \toprule
      Hyperparameters & Default & 3k & 10k & 20k & 50k & 94k \\
      \midrule
      BPE operations & 30k & 5k & 10k & 10k & 12k & 15k \\
      Encoder/decoder layers & 6 & 2 & 2 & 2 & 2 & 2 \\
      Attention heads & 16 & 2 & 2 & 2 & 2 & 2\\
      Embedding dimension & 1024 & 512 & 512 & 512 & 512 & 512 \\
      Feed forward dimension & 4096 & 512 & 1024 & 1024 & 2048 & 2048 \\
      Dropout & 0.3 & 0.3 & 0.3 & 0.3 & 0.3 & 0.3 \\
      Label smoothing & 0.1 & 0.6 & 0.5 & 0.5 & 0.5 & 0.4 \\
      Batch size & 4096 & 4096 & 4096 & 4096 & 4096 & 8192 \\
      \bottomrule
\end{tabular}
%}
\caption{Hyperparameters for optimized transformer. ``Default" denotes the setting of transformer-big. For English-Japanese, BPE operations for ``Vanilla transformer-big" is 40k.} 
\label{hyperp}
 \end{center}
\end{table}

\begin{table}[h]
\begin{center}
%\resizebox{0.9\textwidth}{!}{
\begin{tabular}{ccccccccc}

      \toprule
      \multirow{2}{*}{Model} &\multicolumn{4}{c}{Ja-En}&\multicolumn{4}{c}{En-Ja}\\
      %\midrule
      &  3k &  10k & 20k & 50k & 3k & 10k & 20k & 50k\\
      \midrule
      MASS &8.8&13.8&17.2&21.2&9.1&16.0&20.6&25.0\\
      ENSS &\textbf{11.2}$^\dag$&\textbf{16.7}$^\dag$&\textbf{19.0}$^\dag$&22.1$^\dag$&\textbf{11.7}$^\dag$&\textbf{18.7}$^\dag$&\textbf{22.5}$^\dag$&27.0$^\dag$\\
      \midrule
      BART (text infilling)  & 3.1  & 11.1 & 15.5 & 20.7 & 5.6  & 14.9 & 19.8 & 25.6$^\dag$ \\
      BART + ENSS & 10.7$^\dag$ & 15.8$^\dag$ & 18.9$^\dag$ & \textbf{22.4}$^\dag$ & 10.6$^\dag$ & 17.6$^\dag$ & 21.2$^\dag$ & \textbf{27.2}$^\dag$ \\
      \bottomrule
\end{tabular}
%}
\caption{BLEU scores compared with BART for simulated low/high-resource settings for Japanese--English ASPEC translation using from 3k to 50k parallel sentences for fine-tuning. Results better than MASS with statistical significance $p<0.05$ are marked in \dag.} 
\label{ja-en-bart}
 \end{center}
\end{table}

\begin{table}[h]
\begin{center}
%\resizebox{0.8\textwidth}{!}{
\begin{tabular}{c c c c c c c c c }

      \toprule
     \multirow{2}{*}{Model} &\multicolumn{4}{c}{Ja-Zh}&\multicolumn{4}{c}{Zh-Ja}\\
      %\midrule
      &  3k &  10k & 20k & 50k & 3k & 10k & 20k & 50k \\
      \midrule
      MASS &15.7&20.3&22.4&24.7&19.4&25.9&29.4&32.9\\
      JASS&\textbf{17.1}$^\dag$&\textbf{22.2}$^\dag$&\textbf{23.2}$^\dag$&\textbf{25.2}$^\dag$&21.6$^\dag$&\textbf{27.5}$^\dag$&\textbf{30.4}$^\dag$&\textbf{33.6}$^\dag$\\
      \midrule
      BART (text infilling) & 13.5 & 19.0 & 21.3 & 24.4 & 20.3$^\dag$ & 25.8 & 29.1 & 33.0 \\
      BART + JASS & \textbf{17.1}$^\dag$ & 21.3$^\dag$ & 23.1$^\dag$ & 25.0 & \textbf{21.9}$^\dag$ & \textbf{27.5}$^\dag$ & \textbf{30.4}$^\dag$ & \textbf{33.6}$^\dag$ \\
      \bottomrule
\end{tabular}
%}
\caption{BLEU scores compared with BART for simulated low-resource settings for Japanese--Chinese ASPEC translation using 3k to 50k parallel sentences for fine-tuning. Results better than MASS with statistical significance $p<0.05$ are marked in \dag.}
\label{ja-zh-bart}
 \end{center}
\end{table}

\begin{table}[h]
\begin{center}
%\resizebox{0.8\textwidth}{!}{
\begin{tabular}{c c c c c c c c c}

      \toprule
      \multirow{2}{*}{Model} &\multicolumn{4}{c}{Ja-Zh}&\multicolumn{4}{c}{Zh-Ja}\\
      %\midrule
      &  3k &  10k & 20k & 50k & 3k & 10k & 20k & 50k \\
      \midrule
      MASS & 7.7 &15.4&18.3&23.4&9.6&17.6&23.3&27.1\\
      JASS&\textbf{12.0}$^\dag$&\textbf{17.0}$^\dag$&\textbf{20.1}$^\dag$&\textbf{25.0}$^\dag$&\textbf{16.6}$^\dag$&21.2$^\dag$&\textbf{26.5}$^\dag$&29.2$^\dag$\\
      \midrule
      BART (text infilling) & 5.9  & 14.0 & 18.0 & 21.8 & 8.7  & 17.8 & 24.2$^\dag$ & 28.5$^\dag$ \\
      BART + JASS & 11.4$^\dag$ & 16.5$^\dag$ & 19.4$^\dag$ & 24.3$^\dag$ & 16.2$^\dag$ & \textbf{22.5}$^\dag$ & 26.2$^\dag$ & \textbf{30.0}$^\dag$ \\
      \bottomrule
\end{tabular}
%}
\caption{BLEU scores compared with BART for simulated low-resource settings for Japanese--Chinese Wikipedia translation using from 3k to 50k parallel sentences for fine-tuning. Results better than MASS with statistical significance $p<0.05$ are marked in \dag.} 
\label{ja-zh2-bart}
 \end{center}
\end{table}

\end{document}